\relax
\documentclass[letterpaper]{article} %
\usepackage{aaai20}  %
\usepackage{times}  %
\usepackage{helvet} %
\usepackage{courier}  %
\usepackage[hyphens]{url}  %
\usepackage{graphicx} %
\def\UrlFont{\rm}  %
\usepackage{graphicx}  %
\usepackage[tbtags]{amsmath}
\usepackage{amsthm}
\newtheorem{prop}{Proposition}
\newenvironment{propm}[1]{%
  \renewcommand\theprop{#1}%
  \prop
}{\endprop}
\usepackage[dvipsnames]{xcolor}

\DeclareMathOperator*{\argmin}{arg\,min}
\usepackage{bbm}
\usepackage{amsfonts}
\usepackage{algpseudocode}
\usepackage{algorithm}
\usepackage{subcaption}
\algrenewcommand\algorithmicrequire{\textbf{Input:}}
\algrenewcommand\algorithmicensure{\textbf{Output:}}
\begin{document}
\title{Efficient decorrelation of features using Gramian in Reinforcement Learning}
\author{Borislav Mavrin$^{1,\dagger,\ddagger}$ \hskip 0.5cm Daniel Graves$^{2,\ddagger}$ \hskip 0.5cm Alan Chan$^{3,\dagger,\ddagger}$ \\
$^1$mavrin@ualberta.ca, $^2$daniel.graves@huawei.com, $^3$achan4@ualberta.ca \\
$^\dagger$University of Alberta \hskip 0.3cm $^\ddagger$Noah's Ark Lab Huawei}
\maketitle
\begin{abstract}
Learning good representations is a long standing problem in reinforcement learning (RL).
One of the conventional ways to achieve this goal in the supervised setting is through regularization of the parameters.
Extending some of these ideas to the RL setting has not yielded similar improvements in learning.
In this paper, we develop an online regularization framework for decorrelating features in RL and demonstrate its utility in several test environments.
We prove that the proposed algorithm converges in linear function approximation setting and does not change the main objective of maximizing cumulative reward.
We demonstrate how to scale the approach to deep RL using the Gramian of the features achieving linear computational complexity in the number of features and squared complexity in size of the batch.
We conduct an extensive empirical study of the new approach on Atari 2600 games and show a significant improvement in sample efficiency in 40 out of 49 games.
\end{abstract}

\section{1. Introduction}

Learning a good representation is an important part of machine learning \cite{bengio2013representation}.
In reinforcement learning (RL) and in particular deep RL, achieving a good representation is a significant challenge in learning features that generalize to new states and tasks \cite{farebrother2018generalization}\cite{zhao2019generalization}.
Disentangling factors of variance, especially from highly structured and correlated data such as images, is important to achieving good compact representations \cite{bengio2013representation}.
While some work has been done on studying and improving generalization by regularizing features in RL \cite{farebrother2018generalization}, very little work has been done on disentangling factors of variance in RL in an online setting.
A sensible approach to disentangling, or decorrelating, features in RL is to perform dimensionality reduction techniques like principle component analysis (PCA) on data collected offline \cite{curran2015pca}\cite{liu2015featurebatch}.
Collecting data in advance is a significant disadvantage to these methods; our objective is to demonstrate a theoretically justified approach to decorrelating features in RL in an online manner that is computationally efficient and achieves performance gains, particuarly in deep RL problems.

Decorrelating features in an online manner applicable to RL and computationally efficient is not an easy problem \cite{oja1985stochastic}.
Usually one assumes that the features provided are uncorrelated \cite{bertsekas1996neuro} \cite{parr2008analysis}.
The need for decorrelating features online has increased with the explosion of interest in deep RL.
Firstly, it is often not practically feasible to collect large amounts of data before training the agent especially in infinite or uncountable state spaces.
Secondly, if the environment is non-stationary or extremely large and complex, continual learning and disentangling of the representation can be advantageous to track the important features of the environment for making decisions.

The desire is to learn a generalized representation such that states observed that are similar produce similar value estimates.
In the tabular setting, there is no representation learning; however, in linear and deep function approximation settings, representation learning is an important problem especially in infinitely large state spaces.
Some older works look at state aggregation \cite{moore1991variable} and soft state aggregation \cite{singh1995reinforcement} which involve partitioning the state space according to some property.
While helpful in improving generalization to new states, state aggregation only allows for generalization within a specific partition or cluster.

Neural networks \cite{lecun1988theoretical} are so-called universal approximators, i.e. can approximate any continuous function over a compact set \cite{cybenko1989approximation}, and can be used to achieve generalization over unseen states in deep RL \cite{zhang2016understanding} \cite{NIPS2017_7176} \cite{shwartz2017opening}.
Other methods of learning generalizable feature representations include kernel methods \cite{boser1992training}.
We focus on decorrelating linear and deep representations in RL.
On supervised problems, decorrelating features has been shown to improve generalization.
\cite{oja1982simplified} is one of the earliest results of applying decorrelation in neural networks. 
The decorrelating of features was hypothesized to explain, in part, the underlying working principle of the visual cortex V1 \cite{NIPS2009_3868}. 
In more recent research, it was empirically demonstrated that feature decorrelation improves generalization and reduces overfitting in a variety of tasks and across different models \cite{cheung2014discovering} \cite{cogswell2015reducing} \cite{chang2018scalable}.
\cite{cogswell2015reducing} show experimentally that decorrelating features is competitive in performance to dropout \cite{srivastava2014dropout}, a widely used regularizer approach, and can be combined with dropout to achieve even better performance.

Unfortunately, common methods of regularizing a network in supervised learning seems to improve generalization of RL but does not improve performance \cite{farebrother2018generalization}.
The authors performed an empirical study on the effects of dropout and $L_2$ weight regularization on the generalization properties of DQN \cite{mnih2015human} and concluded that they help learn more general purpose representations.
Another approach proposed an $L_2$ weight regularization approach in \cite{farahmand2009regularized} focusing on the supporting theory and lacking experimental results on the generalization properies of the method.
While generalizing to similar tasks in RL is an important problem, our focus is on a theoretically grounded approach to decorrelating features while the agent interacts online with the environment and measure its impact on learning performance.
Earlier work by \cite{mavrin2019deep} introduced online feature decorrelation in RL and provided only empirical results.
Their work was computationally inefficient in the number of features since it scaled quadratically with the number of features while the proposed approach scales linearly with the number of features with non-linear function approximation. 

The main contributions of this paper are as follows:
\begin{itemize}
\item Develop an online algorithm for decorrelating features in RL
\item Prove its convergence
\item Justify theoretically that the proposed regularizing decorrelation loss in the RL setting does not change the RL objective
\item Scale up linearly in features in the deep RL setting and demonstrate empirically that decorrelating features in RL improves performance on 40 out of 49 Atari 2600 games
\end{itemize}

The rest of the paper is organized as follows.
In section 2, we present the proposed algorithm and justify it theoretically.
In section 3, we show empirically that the proposed algorithm for online decorrelation of features provides performance benefits on many RL problems.
In section 4, we draw some conclusions and in the appendix we show the proofs and derivations for the algorithm along with further details on the experiments.

\section{Decorrelating features in RL}
In this section we first describe the problem setting and notation.
Then, we introduce the decorrelating constraint into the Mean Squared Value Error minimization problem with TD(0) \cite{sutton1988learning}.
We consider the linear function approximation case in our theoretical analysis due to its tractability, and we show that the adding a regularizer term in this case does not change the original TD(0) solution.
Finally, we prove the convergence of the stochastic approximation algorithm for TD(0) with the decorrelating regularizer by applying the Borkar-Meyn theorem [chapter 2 of \cite{borkar2009stochastic}].

\subsection{Problem setting}
We model the environment as a Markov Decision Process with state transition probabilities $P(s'|s,a)$ and initial state distribution $\mu: S \to [0, 1]$ invariant with respect to $P$, i.e. $P\mu=\mu$  and $D=diag\{\mu(s_1), \dots \mu(s_n)\}$.
We also assume finite state and finite actions spaces, i.e. $S=\{s_1, \dots s_n \}$, $\mathcal{A}=\{a_1, \dots a_m \}$, with a given set of feature functions $\phi: S \to \mathbb{R}^d$ and $\Phi=[\phi(s_1), \dots \phi(s_n)]^T$. We denote the set of all terminal states by $\mathcal{T}$.
The reward is a real-valued function $R: S \times \mathcal{A} \to \mathbb{R}$.
Agent's objective is to maximize expected discounted sum of rewards: $\mathbb{E}_\pi [\sum_t \gamma^t R_t | S_t=s, A_t=a]$ by adjusting its decision rule represented by a policy function $\pi: S \to \mathcal{A}$. Specifically, in Q-learning, the agent estimates the state-action values $Q^\pi(s,a) = \mathbb{E}_\pi [\sum_t \gamma^t R_t | S_t=s, A_t=a]$ and given $(s,a) \in S \times \mathcal{A}$ picks an action with the highest state-action value.
In the function approximation case, state-action values are approximated by a linear function: $Q^\pi(s,a) \approx \phi(s,a)^T \theta$, where $\theta \in \mathbb{R}^d$.
The feature function $\phi(s,a)$ can be freely chosen, including for example a one-hot state encoding to a Neural Network.

\subsection{Decorrelating regularizer and analytical gradients}
Features are decorrelated when the covariance matrix of the state features is diagonal, i.e. given a batch of states features $\Phi = [\phi(s_1) \dots\phi(s_n)]^T$, the off-diagonal elements that correspond to the covariances vanish $\forall i<j \hskip 0.3cm e_i \Phi^T \Phi e_j = 0$, where $e_1 = [1, 0 \dots 0]^T, \dots e_n = [0, \dots 0, 1]$ are the standard basis vectors.
In the case when features are correlated it is possible to find a transformation of features, $A$ via diagonalization or SVD, s.t. $\forall i<j \hskip 0.3cm e_i A^T \Phi^T \Phi A e_j = 0$
\cite{jolliffe1986principal}.
Such a solution requires access to the features of all states. In the case of decorrelating features in RL, it is desirable to learn such a transformation $A$ online.
One possible loss function for learning the value function of a given policy is to augment the Mean Squared Value Error (MSVE) loss,
\begin{equation} \label{eq:full_loss}
\begin{split}
	L_{TD}(\theta) = 0.5 \sum_s \mu(s)[R(s) + \gamma \phi(s')^T \theta - \phi(s)^T \theta]^2,
\end{split}
\end{equation}
with a feature decorrelating regularizer, which is the L2 loss of the off-diagonal elements of the $A^T \Phi^T \Phi A$.
\begin{equation} \label{eq:full_aug_loss}
\begin{split}
	& L_{REG}(\theta, A) \\
	& = 0.5 \sum_s \mu(s)[R(s) + \gamma \phi(s')^T A \theta - \phi(s)^T A \theta]^2 \\
 	& + 0.5 \lambda \sum_{i<j} [e_i\sum_s (\mu(s) A^T \phi(s) \phi(s)^T A) e_j]^2.
\end{split}
\end{equation}

\noindent Geometrically, orthogonal transformation $A$ of features $\phi$ can be viewed as a rotation (given $det(A)=1$) of the feature space. Therefore, if the feature space is rotated, then the value function defined over that space does not change if its parameters are adjusted accordingly. However, this form of the loss is not as simple as it could be because of the two sets of parameters $(\theta, A)$.

Note that matrix diagonalization is unique only if we restrict the diagonalizing matrix $A$ to be orthogonal. If the column vectors of the diagonalizing matrix are rescaled, i.e. $\tilde{A} = [\theta_1 a_1 \dots \theta_d a_d]$, the resulting matrix, while no longer guaranteed to be orthogonal, is still a diagonalizing matrix. This observation allows to reduce the number of parameters in Equation \eqref{eq:full_aug_loss} from $(\theta, A)$ to $A$ since the regularization term will be zero for any such rescaling of a matrix $A$ into $\tilde{A}$.

The benefit of such a reduction is two-fold. First, fewer parameters may lead to faster learning. Second, theoretical analysis of the reduced problem is simpler. Observe that the original problem in Equation \eqref{eq:full_aug_loss} is parametrized by a vector $\theta$ and a matrix $A$ which belong to different metric spaces. On the other hand, the reduced problem is parametrized by a single matrix $A$ and the update is easier to analyze.

The loss of the reduced problem is
\begin{equation} \label{eq:red_aug_loss}
\begin{split}
	& L_{REG}(A) \\
	& = 0.5 \sum_s \mu(s)[R(s) + \gamma \phi(s')^T A \mathbbm{1}  - \phi(s)^T A \mathbbm{1}]^2 \\
 	& + 0.5 \lambda \sum_{i<j} [e_i\sum_s (\mu(s) A^T \phi(s) \phi(s)^T A) e_j]^2
\end{split}
\end{equation}

The following proposition ensures that solving Equation \eqref{eq:red_aug_loss} also solves Equation \eqref{eq:full_loss}.

\begin{prop}\label{prop:param_loss_td_loss_equivalence}
	Let $A$ be a global minimum of Equation \eqref{eq:red_aug_loss}. Then $A^T \Phi^T D \Phi A$ is a diagonal matrix.
	Furthermore, the global minimum values of Equation \eqref{eq:red_aug_loss} and Equation \eqref{eq:full_loss} are equal.
\end{prop}
\begin{proof}
	See Appendix.
\end{proof}

\subsection{Convergence}
In order to obtain an online algorithm one needs the semi-gradient of the Equation \eqref{eq:red_aug_loss}.
The reparametrized update is \footnote{See Appendix for derivation.}:
\begin{equation} \label{eq:red_a_update}
\begin{split}
	&\frac{ \partial L_{REG}(A)}{ \partial A} = \\
	&-\sum_s \mu(s) \phi(s) \mathbbm{1}^T [R(s) + \gamma \phi(s')^T A \mathbbm{1} - \phi(s)^T A \mathbbm{1}]\\
	&+ \lambda \Phi^T D \Phi A [E_1^T \tilde{D} E_2 + E_2^T \tilde{D} E_1]
\end{split}
\end{equation}
where for $\phi \in \mathbb{R}^d$
\begin{itemize}
	\item $E_\mathbbm{1}^T =[e_1 \mathbbm{1}^T_{d-1},  e_2 \mathbbm{1}^T_{d-2} \dots e_{d-1}]$
	\item $E_2^T=[e_2 \mathbbm{1}^T,  e_3 \mathbbm{1}^T_2 \dots e_d \mathbbm{1}^T_{d-1}]$
	\item $\tilde{D} = diag\{\tilde{\sigma}_{ij}\}_{i<j}$ a diagonal matrix constructed from the off diagonal elements of the covariance matrix for the transformed features, i.e. $\tilde{\sigma} = A^T \Phi^T \Phi A$
	\item $\{e_i\}_i$ are standard basis vectors of $\mathbb{R}^d$
	\item $\mathbbm{1}_{n} \in \mathbb{R}^n$ is the vector of ones
\end{itemize}
For example in $\mathbb{R}^3$ $E_1^T = \{e_1, e_1, e_2 \}$, $E_2^T = \{e_2, e_3, e_3 \}$, $\tilde{D} = diag\{\tilde{\sigma}_{12}, \tilde{\sigma}_{13}, \tilde{\sigma}_{23}\}$.

In order to apply the Borkar-Meyn theorem, we have to ensure the iterates are bounded; i.e. $sup_n ||A_n|| < \infty$. One option is to simply assume this condition outright, which is often done in practice. In that case we arrive at the following convergence result.

\begin{prop}\label{prop:borkar_convergence}
Assuming that $\sup_n \|A_n \| < \infty$ almost surely and the Robbins-Munro step-size conditions, the iterates of the stochastic update converge to a compact, connected, internally chain-transitive set of $\dot{A}(t) = h(A(t))$. Additionally, if $\Phi$ is of full rank, there exists such a set that contains at least one equilibrium of $\dot{A}(t) = h(A(t))$.

\end{prop}
\begin{proof}
	See Appendix.
\end{proof}

Alternatively, a more theoretically sound approach is to ensure $\sup_n \|A_n \| < \infty$. We achieve this by introducing an additional step that projects\footnote{See section 5.4 of \cite{borkar2009stochastic}.} $A$ onto the space of orthogonal matrices i.e. $\tilde{A} = \argmin_{Q^TQ=I} ||A-Q ||_F$, where $||.||_F$ is the Frobenius norm. The solution to this projection problem is provided by the following proposition.

\begin{prop} \label{prop:bound}
The projection of a square matrix $A_{n \times n}$ onto the space of orthogonal matrices $\mathcal{Q}$ is given by $UV^T$, where $U$ and $V$ are obtained from SVD of $A=U\Sigma V^T$.
\end{prop}
\begin{proof}
	See Appendix.
\end{proof}

\noindent For the Borkar-Meyn theorem, we also need to show that the semi-gradient $h(A)$ above is Lipschitz over its domain. If we project the iterates to $\mathcal{Q}$, the following proposition gives us the Lipschitz property. %
\begin{prop} \label{prop:lipschitz}
	$h(A)=\frac{ \partial L_{REG}(A)}{ \partial A}$ defined by Equation \eqref{eq:red_a_update} over the space of orthogonal matrices satisfies the Lipschitz condition.
\end{prop}
\begin{proof}
	See Appendix.
\end{proof}
Lastly, Borkar-Meyn theorem conditions on the Martingale difference noise also follow from the projection step. 
The proof of this result is in the Proposition \ref{prop:martingale_diff} in the Appendix. 

%However, the only missing condition that prevents us from immediately applying Borkar-Meyn theorem is the condition on the boundary of $\mathcal{Q}$.
%It would be interesting to investigate this issue further.

\subsection{Linear Q-learning}
Incorporating the decorrelating update into Q-learning in the linear setting is quite straightforward. The only change is in the weight update step.

    \begin{algorithm}
      \caption{linear Q-learning with feature decorrelation} \label{alg:q_decor_red}
      \begin{algorithmic}[1]
        \Require $\epsilon \in [0, 1]$, $A \in \mathbb{R}^{d \times d}, \gamma \in [0, 1)$ %
        \Require $\alpha > 0$, $\lambda > 0$
        \State $s \sim \mu(s)$
        \State $a' \leftarrow \arg \max_a \phi(s,a)^T A \mathbbm{1}$
        \While{$s'$ is not terminal}
          \State take action $a'$ with probability $1-\epsilon$ or a random action with probability $\epsilon$
          \State observe $r,s'$
          \State Update $A$: $A \leftarrow A + \phi(s,a) \mathbbm{1}^T$
          \Statex $\big [r + I_{\{s' \not\in \mathcal{T}\}} \gamma \max_a \phi(s',a)^TA \mathbbm{1} - \phi(s,a)^T A \mathbbm{1} \big]$
          \Statex $ - \lambda \phi(s,a) D \phi(s,a)^T A [E_1^T \tilde{D} E_2 + E_2^T \tilde{D} E_1]$
          \State $a' \leftarrow \arg \max_a \phi(s',a)^T A \mathbbm{1}$
        \EndWhile
      \end{algorithmic}
    \end{algorithm}
In practice we found $A$ to be close to orthogonal without projection step step in Algorithm \ref{alg:q_decor_red}; in this case, the complexity reduces to quadratic in the number of features. 
Such complexity might be still prohibitive when scaling to higher dimensional spaces.

\subsection{Scaling up to deep RL}
In the case of high dimensional-representations such as with Neural Networks, squared complexity in the features in Equation \eqref{alg:q_decor_red} might be a significant computational limitation.
One possible solution to the problem was suggested by \cite{NIPS2009_3868}.
The main idea is to move the squared complexity in features to the sample size by representing the covariance by a Gramian matrix.
Such an approach is based on the fact that in practice neural networks are trained with Mini-Batch SGD which assumes a fixed, small batch size.

For completeness, we reproduce the result by \cite{NIPS2009_3868} in notation consistent with our work. Let $Cov = \Phi^T \Phi$, where $\phi$ is the last hidden layer of the NN. Then the sum of squared elements of $Cov$ is
\begin{equation}
\begin{split}
 	 & \sum_{i,j} Cov_{i,j}^2 = Tr(\Phi^T \Phi \Phi^T \Phi) \\
 	 &  = Tr(\Phi \Phi^T \Phi \Phi^T) = \sum_{i,j} G_{i,j}^2
\end{split}
\end{equation}
where $G=\Phi \Phi^T$. Therefore, the decorrelating regularizer is equivalent to
\begin{equation}
\begin{split}
 	 & \sum_{i \ne j} Cov_{i,j}^2 = \sum_{i,j} Cov_{i,j}^2 - \sum_{i=j} Cov_{i,j}^2 \\
 	 & = Tr(\Phi^T \Phi \Phi^T \Phi) - \sum_{i=j} Cov_{i,j}^2 \\
 	 & = Tr(\Phi \Phi^T \Phi \Phi^T) - \sum_{i=j} Cov_{i,j}^2 \\
 	 & = \sum_{i,j} G_{i,j}^2 - \sum_{i} Var_{i}^2
\end{split}
\end{equation}
We extend this idea to the deep RL setting via the following optimization objective:
\begin{equation} \label{eq:dqn-gram-loss}
	L_{\text{DQN-Gram}}=L_{\text{DQN}}+\lambda \big [ \sum_{i,j} G_{i,j}^2 - \sum_{i} Var_{i}^2 \big ]
\end{equation}
In Equation \eqref{eq:dqn-gram-loss} the complexity is linear in features and squared in the sample size which allows to scale Algorithm \ref{alg:q_decor_red} to the deep RL setting more efficiently compared to \cite{mavrin2019deep}. We call our algorithm DQN-Gram. It is outlined in Algorithm \ref{alg:dqn_gram}.

    \begin{algorithm}
      \caption{DQN-Gram}\label{alg:dqn_gram}
      \begin{algorithmic}[1]
        \Require $\lambda > 0, \gamma \in [0, 1$
        \State Initialize $\phi(s|w)$, target $\hat{\phi}(s|w^-)$
        \State Initialize $\{\theta_a\}_{a \in \mathcal{A}}$, target $\{\hat{\theta}^-_a\}_{ \in \mathcal{A}}$
        \State Initialize replay buffer $\mathcal{D}$
        \State $s \sim \mu(s)$
        \State $a' \leftarrow \arg \max_a \phi(s|w)^T \theta_a$
        \While{$s'$ is not terminal}
          \State Take action $a'$ observe $r,s'$
          \State Store transition $(s,a,r,s')$ in $\mathcal{D}$
          \State Sample mini batch $\{(s,a,r,s')_n\} \sim \mathcal{D}$
          \State Update $w,\{\theta_a\}_{a \in \mathcal{A}}$ by gradient descent on
          \Statex $\sum_n [r_n + I_{\{s_n \not\in \mathcal{T}\}} \gamma \max_a \hat{\phi}(s'_n|w^-)^T\hat{\theta}^-_a - \phi(s_n|w)^T\theta_{a_n}]^2  + \lambda \sum_{i,j} G_{i,j}^2 - \sum_{i} Var_{i}^2$
          \State $a'=\arg \max_a \phi(s'|w)^T \theta_a$
        \State Reset $w^- \leftarrow w$ and $\theta^-_a \leftarrow \theta_a$ every $C$ steps
        \EndWhile
      \end{algorithmic}
    \end{algorithm}

\section{Experiments}
We perform extensive experiments of our algorithm in a few settings: stochastic linear regression (SLR),
linear RL and deep RL.
Experiments reveal significant improvement in sample efficiency.
We also examine the properties of the trained neural network and conclude that decorrelating regularizer allows to increase model capacity and the number of the learned features.

\subsection{Stochastic Linear Regression}
Decorrelating objective developed in the previous section applies to the SLR case, since it is a special case of \eqref{eq:full_aug_loss} where the target is a fixed label $y$ and $\phi=x$.
The data $x \in \mathbb{R}^2$ for the SLR is generated as follows: $x \sim N(0, \Sigma)$, label $y = x_1 + 2 x_2$. 
We perform two sets of experiments: 
\begin{itemize}
 \item data is uncorrelated, i.e. $\Sigma=I$; 
 \item data is correlated $ \Sigma = 
\begin{bmatrix}
1. & 0.99 \\
0.99 & 1.
\end{bmatrix}
$.
\end{itemize}

\begin{figure}[th]
    \centering
    \begin{subfigure}{0.45\linewidth}
        \includegraphics[width=0.99\textwidth]{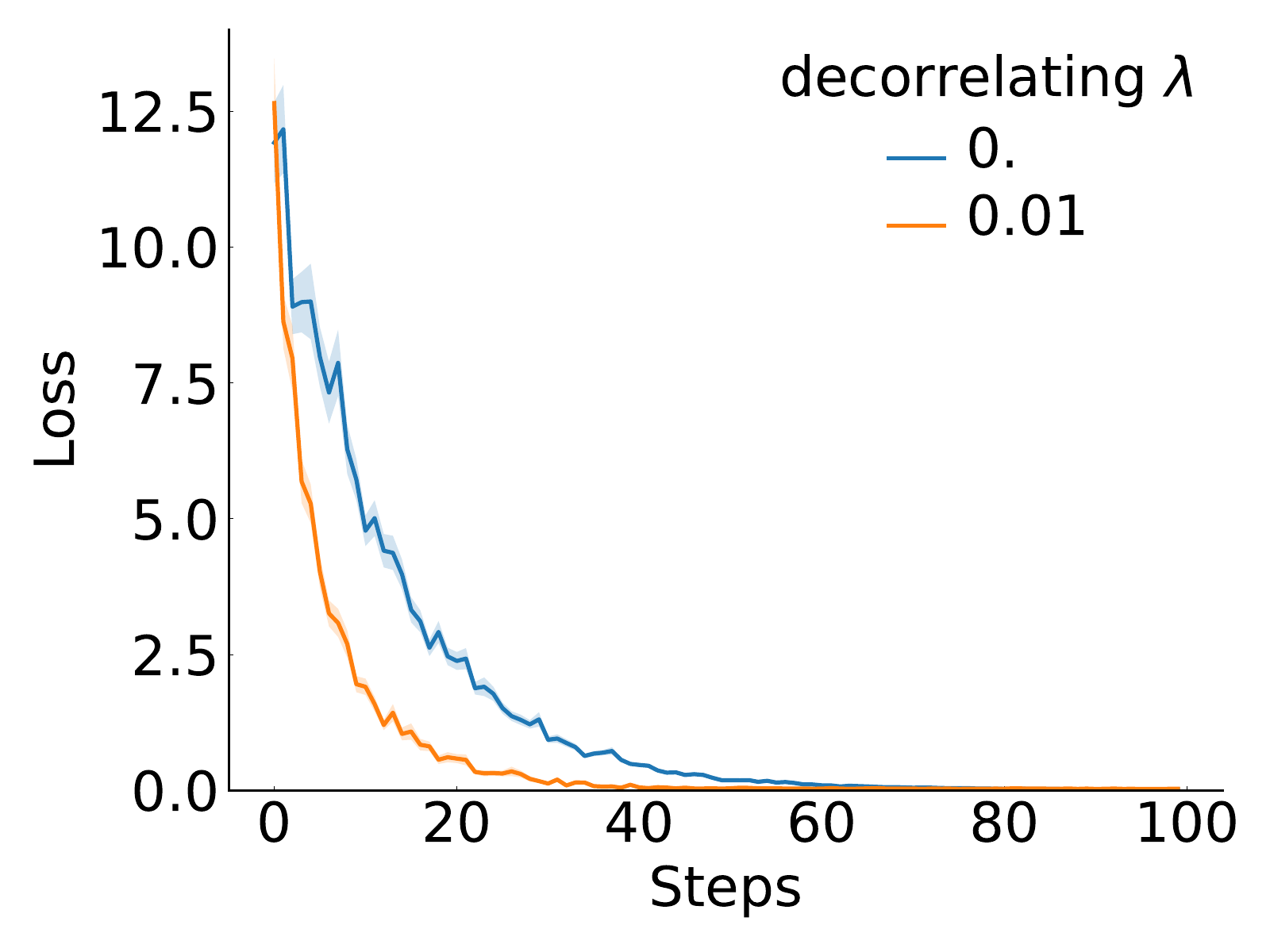}
        \caption{Correlated data.}
    \end{subfigure}\qquad
	\begin{subfigure}{0.45\linewidth}
        \includegraphics[width=0.99\textwidth]{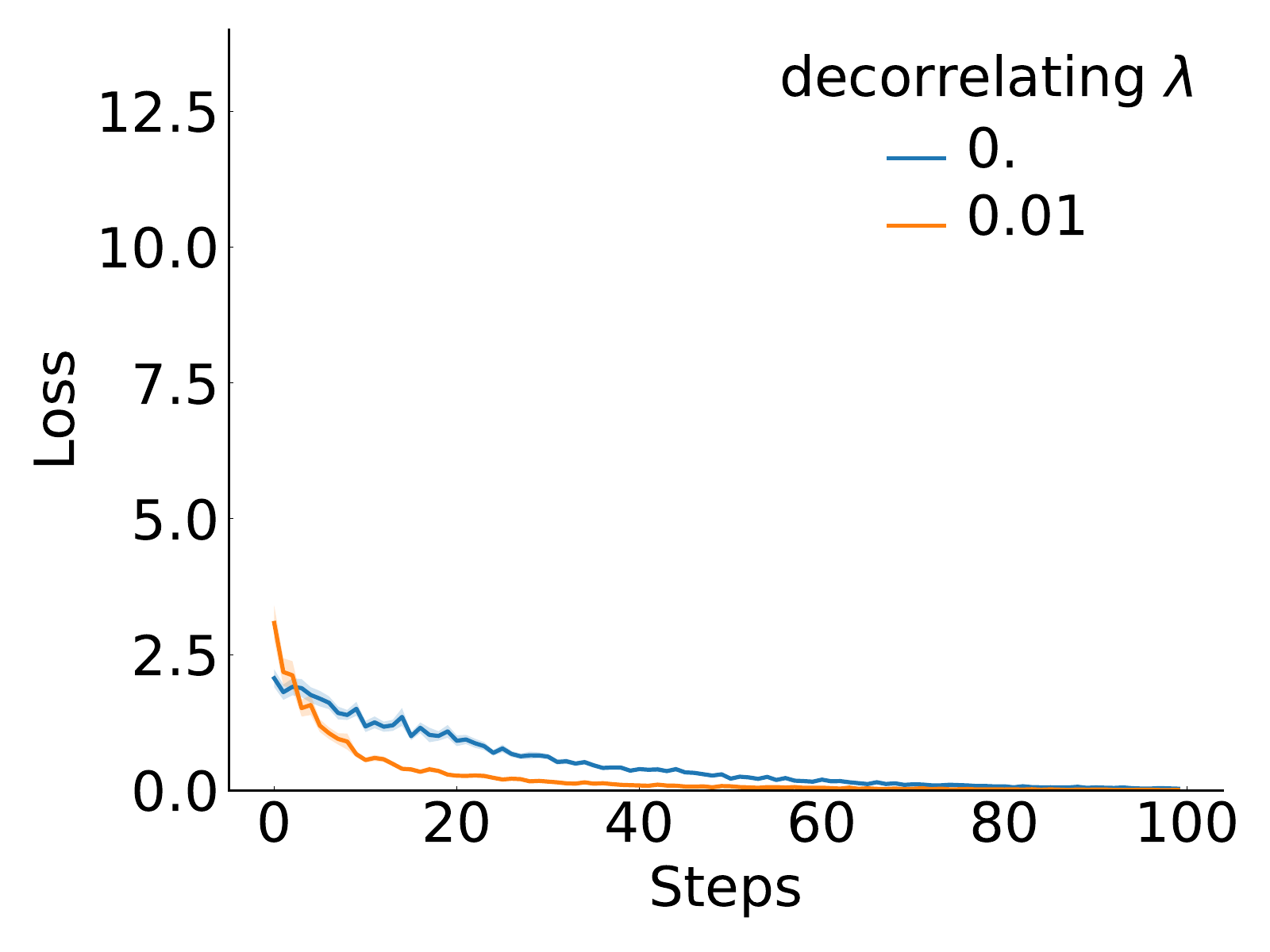}
        \caption{Uncorrelated data.}
    \end{subfigure}
    \caption{Training curves for Stochastic Linear Regression. Results are averaged over 1,000 runs. Shaded area represents std err.}
    \label{fig:slr}
    \end{figure}

We set the learning rate to 0.01 and mini-batch size to 1. The results are reported in Figures \ref{fig:slr}. 
In both cases SLR with decorrelating regularizer has better sample efficiency.

\subsection{Linear RL}
\begin{figure}[th]
    \centering
    \begin{subfigure}{.45\linewidth}
        \includegraphics[width=0.99\textwidth]{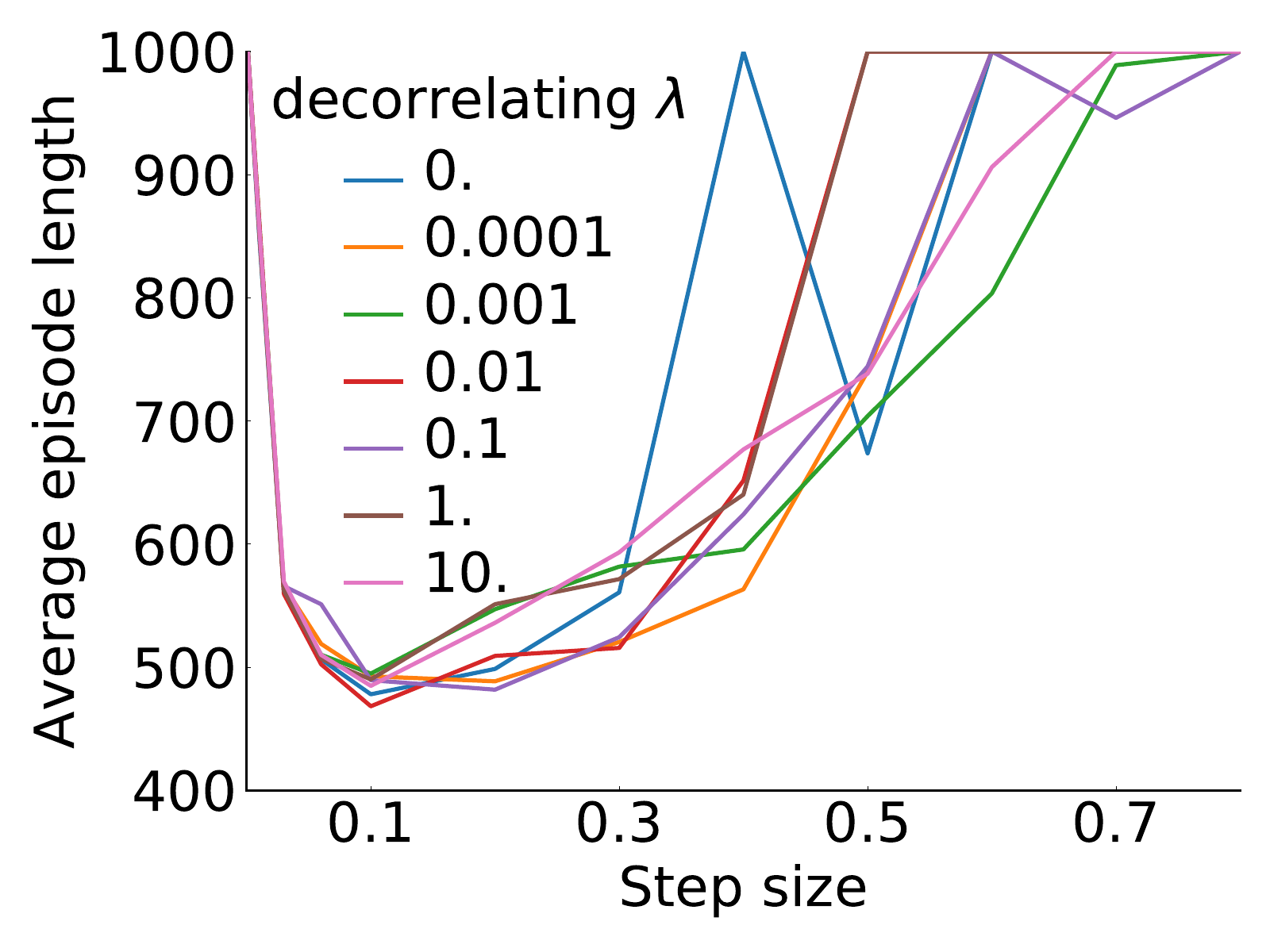}
        \caption{Features with higher correlation.}
    \end{subfigure}
    \begin{subfigure}{.45\linewidth}
        \includegraphics[width=0.99\textwidth]{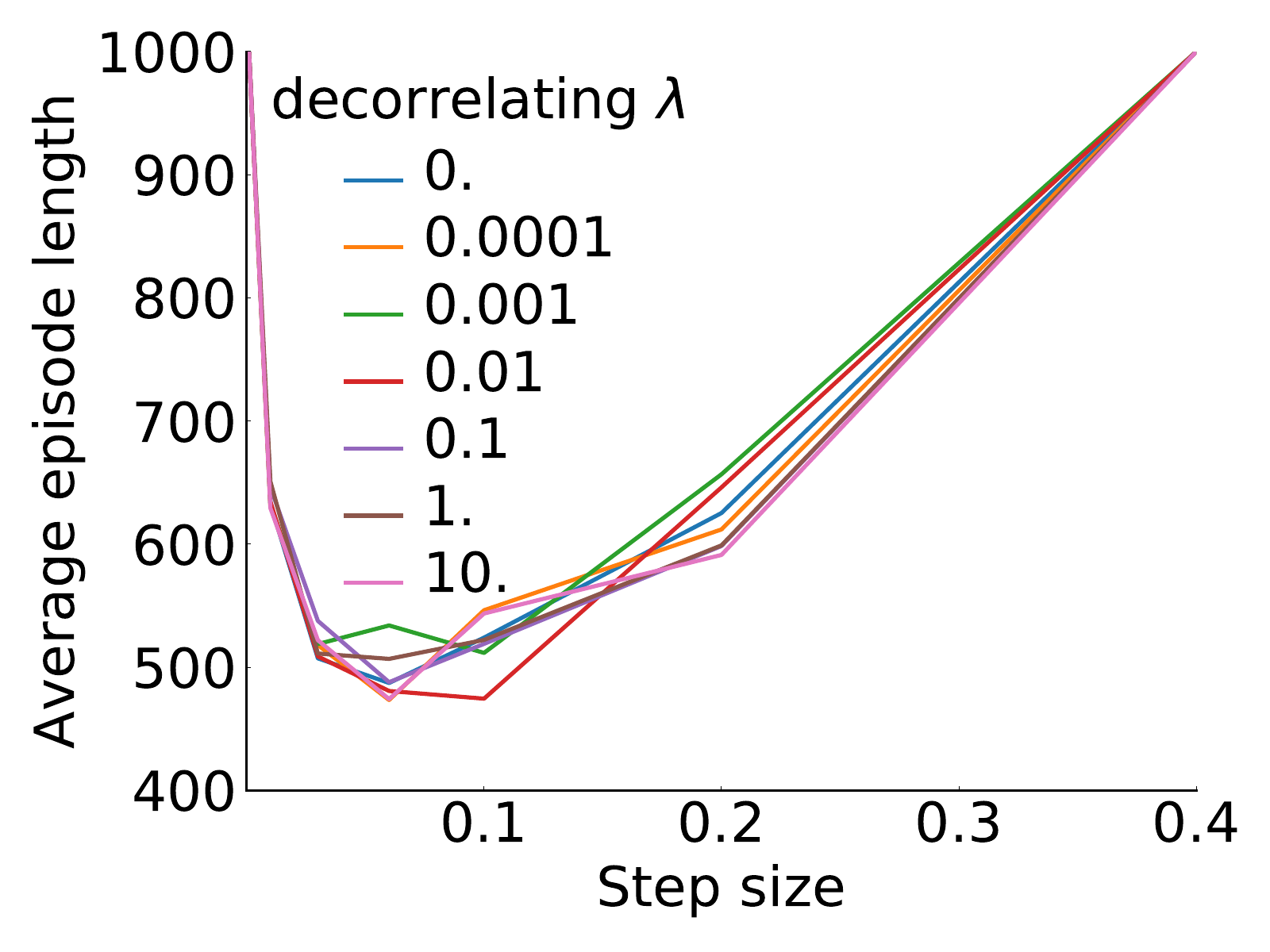}
        \caption{Features with lower correlation.}
    \end{subfigure}
    \caption{Mountain car with tile coding (2 tiles). Episode length averaged over first 50 episodes 100 runs. Standard errors are thinner than the presented curves.}
    \label{fig:mc}
    \end{figure}

For the linear RL setting we perform experiments on the Mountain car environment \cite{sutton1996generalization}. 
We use tile coding as a feature approximator with 2 tiles.
Note that tile coding produces sparse features with low correlation.
In order to control for correlation we augment the feature space by adding extra dimensions that are copies of the original features.
For example if $\phi \in \mathbb{R}^d$, then $\tilde{\phi} = [\phi_1, \dots \phi_d, \phi_i] \in \mathbb{R}^{d+1}$ with $i \in \{1, \dots d \}$.
To test the hypothesis of robustness with respect to correlated features we perform a step size sweep jointly with the $\lambda$ sweep.
For each (step-size, $\lambda$) pair we average the number of steps in each episode over 50 episodes and 100 runs [following (Sutton 2018 p 248)].
It can be concluded from the results presented in Figure \ref{fig:mc} that decorrelating Q-learning improves performance and decreases parameter sensitivity when the features have relatively high correlation, compared to when the features already have relatively low correlation.

\subsection{Deep RL}
\begin{figure}[h!]
	\centering
	\includegraphics[width=0.47\textwidth]{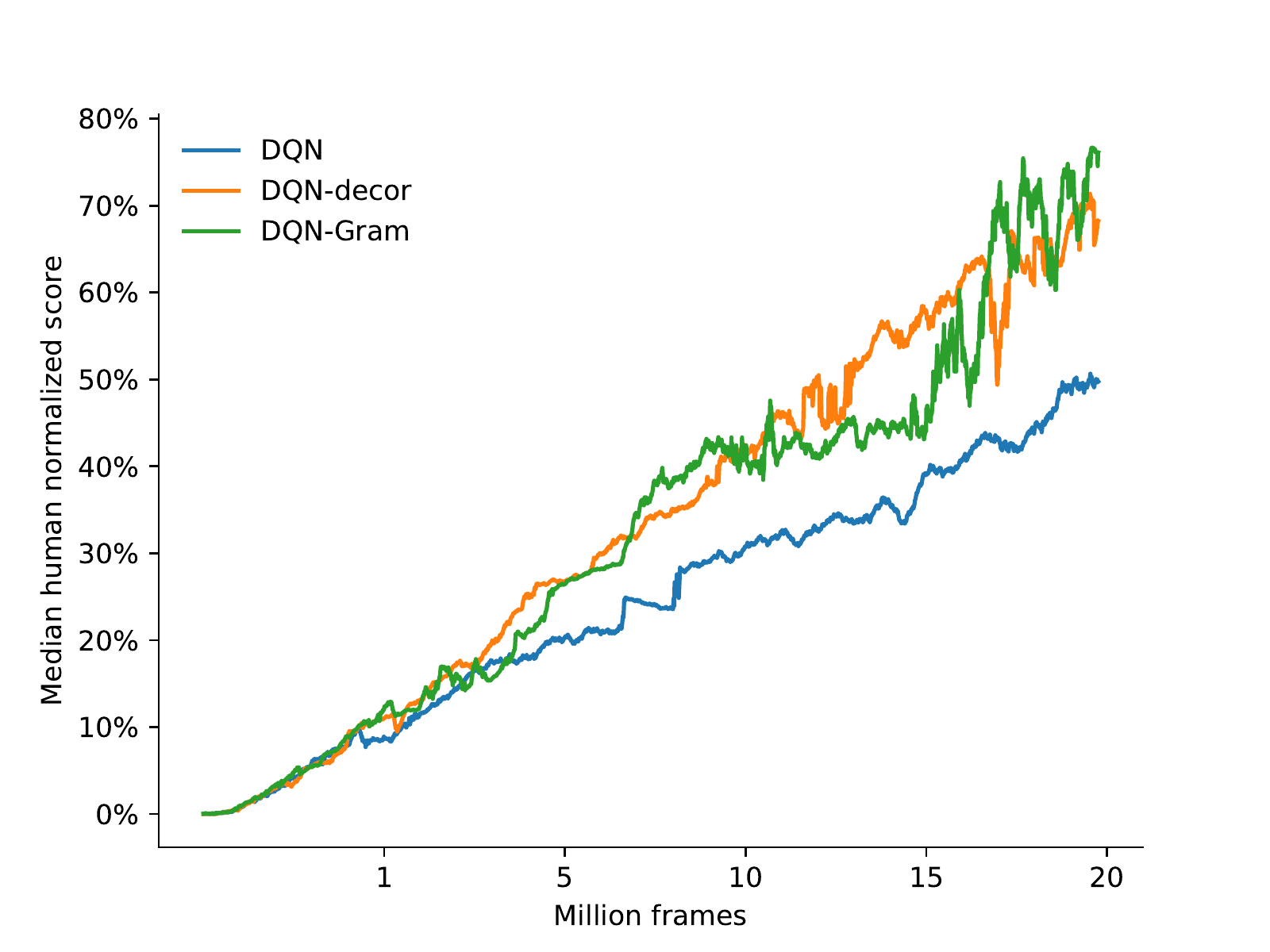}
    \caption{Median human normalized scores across 49 Atari 2600 games.}
    \label{fig:agg_normed}
\end{figure}

\begin{figure*}[ht!]
	\centering
	\includegraphics[width=0.99\textwidth]{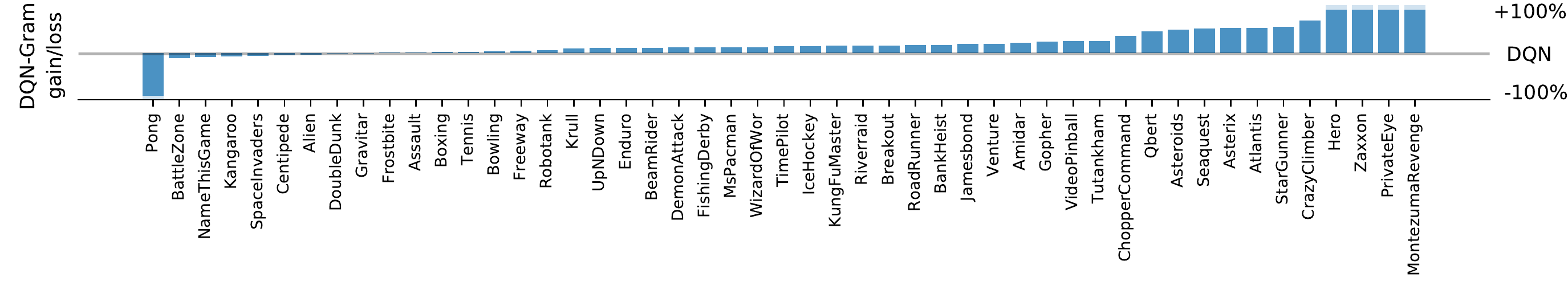}
    \caption{DQN-gram performance gain/loss over DQN in area under the learning curve for 49 Atari 2600 games.}
    \label{fig:auc}
\end{figure*}

We investigate decorrelating features of the value function in deep RL with DQN.
In our experiments we set all the hyperparameters following \cite{mnih2015human} except for the optimizer and the learning rate. In our setup we used Adam \cite{kingma2014adam} with 1e-4 as the learning rate.
It can be seen from Figure \ref{fig:agg_normed} that DQN-Gram and DQN-decor produce equivalent simulation results as predicted by their theoretical equivalence. However, DQN-Gram has lower computational feature complexity.

    \begin{figure}[th!]
    \centering
    \begin{subfigure}{.45\linewidth}
        \includegraphics[width=0.99\textwidth]{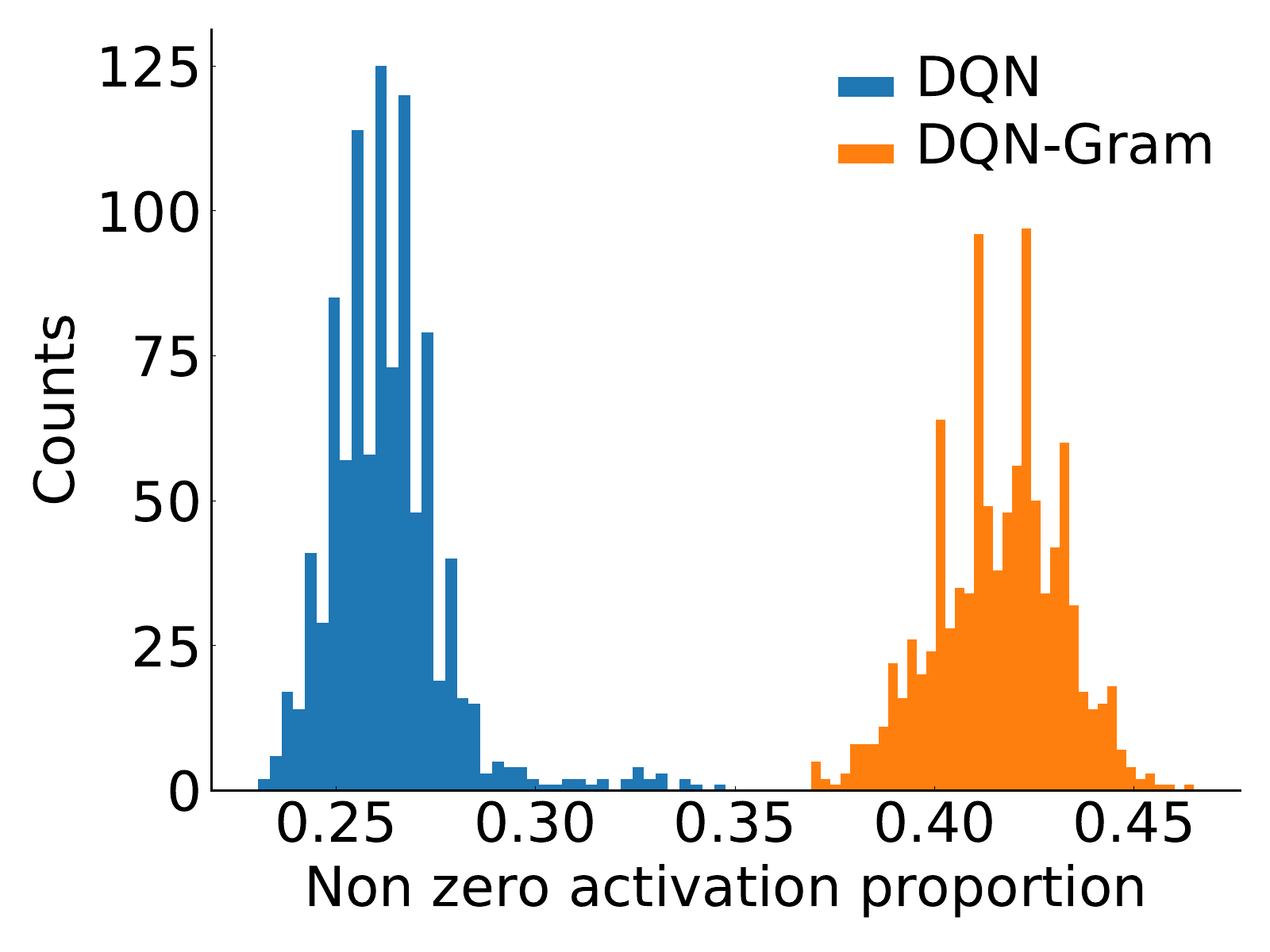}
        \caption{Nonzero activation histogram for DQN and DQN-Gram.}
    \end{subfigure}\qquad
    \begin{subfigure}{.45\linewidth}
        \includegraphics[width=0.99\textwidth]{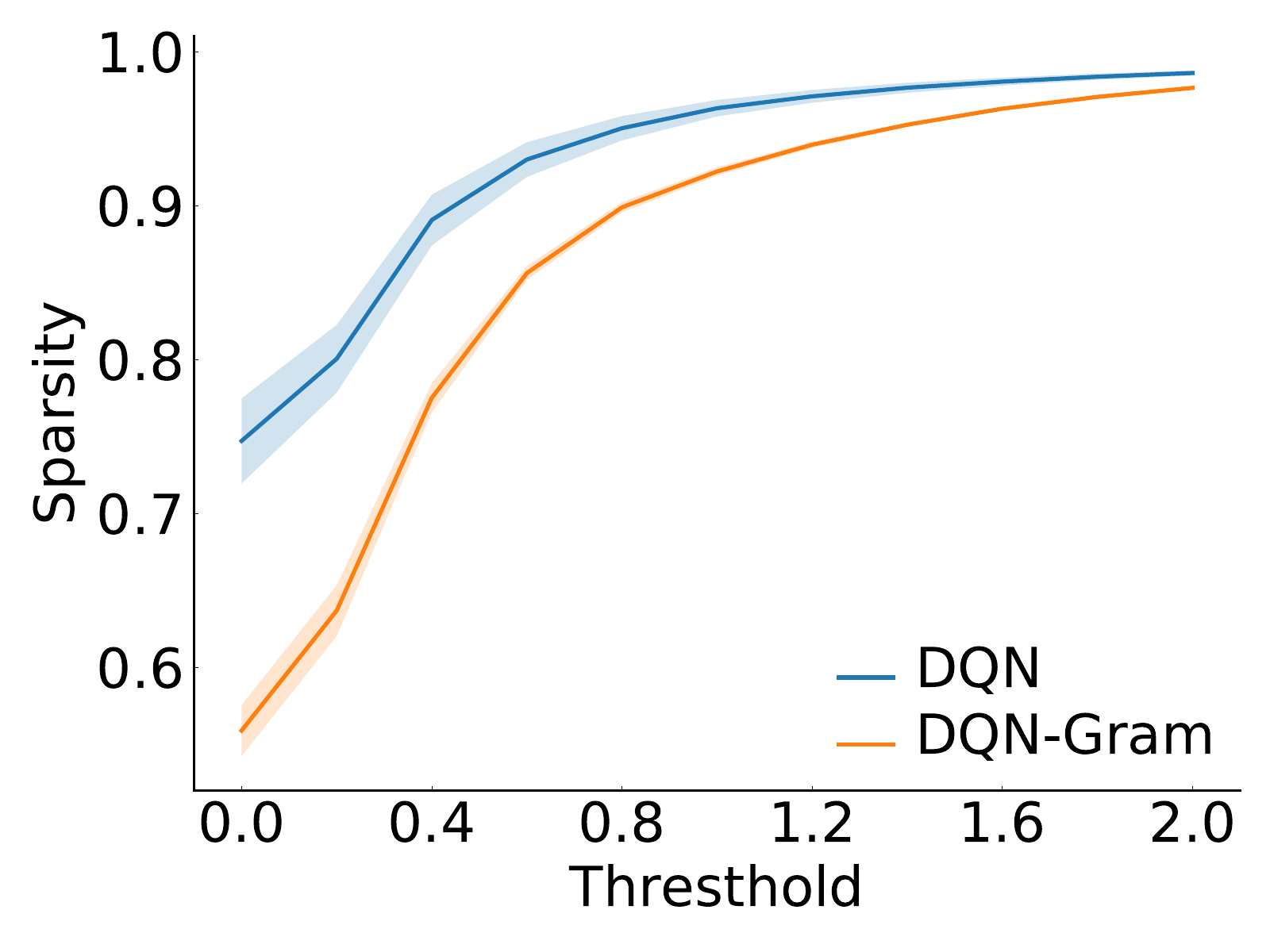}
        \caption{Sparsity of features with varying threshold for 3 seeds. Bars represent std error.}
    \end{subfigure}
    \caption{Feature statistics for Seaquest Atari 2600.}
    \label{fig:sparsity}
    \end{figure}

In order to get a better understanding of the improved performance of the model with the decorrelating regularizer we study the properties of the neural networks in question. 

We hypothesize that there is difference in sparsity of the learned features between DQN and DQN trained with the decorrelating regularizer. In order to test this hypothesis we compute the histogram of non zero activations of features (last hidden layer) of the trained agents across 1 million states. In addition we measure the sparsity of activations by varying the threshold, i.e. 
\begin{equation}
\text{sparsity}_\epsilon = 1 - \frac{\sum_{i=1}^N{I_{|\phi_i| > \epsilon}}}{N}
\end{equation}
It can be seen from Figure \ref{fig:sparsity} that there is significant difference in sparsity. Interestingly, the model trained with the decorrelating regularizer has more dense representations indicating a richer representation that potentially better exploits the representational capabilities of the network architecture.

    \begin{figure}[th!]
    \centering
    \begin{subfigure}{0.45\linewidth}
        \includegraphics[width=0.9\textwidth]{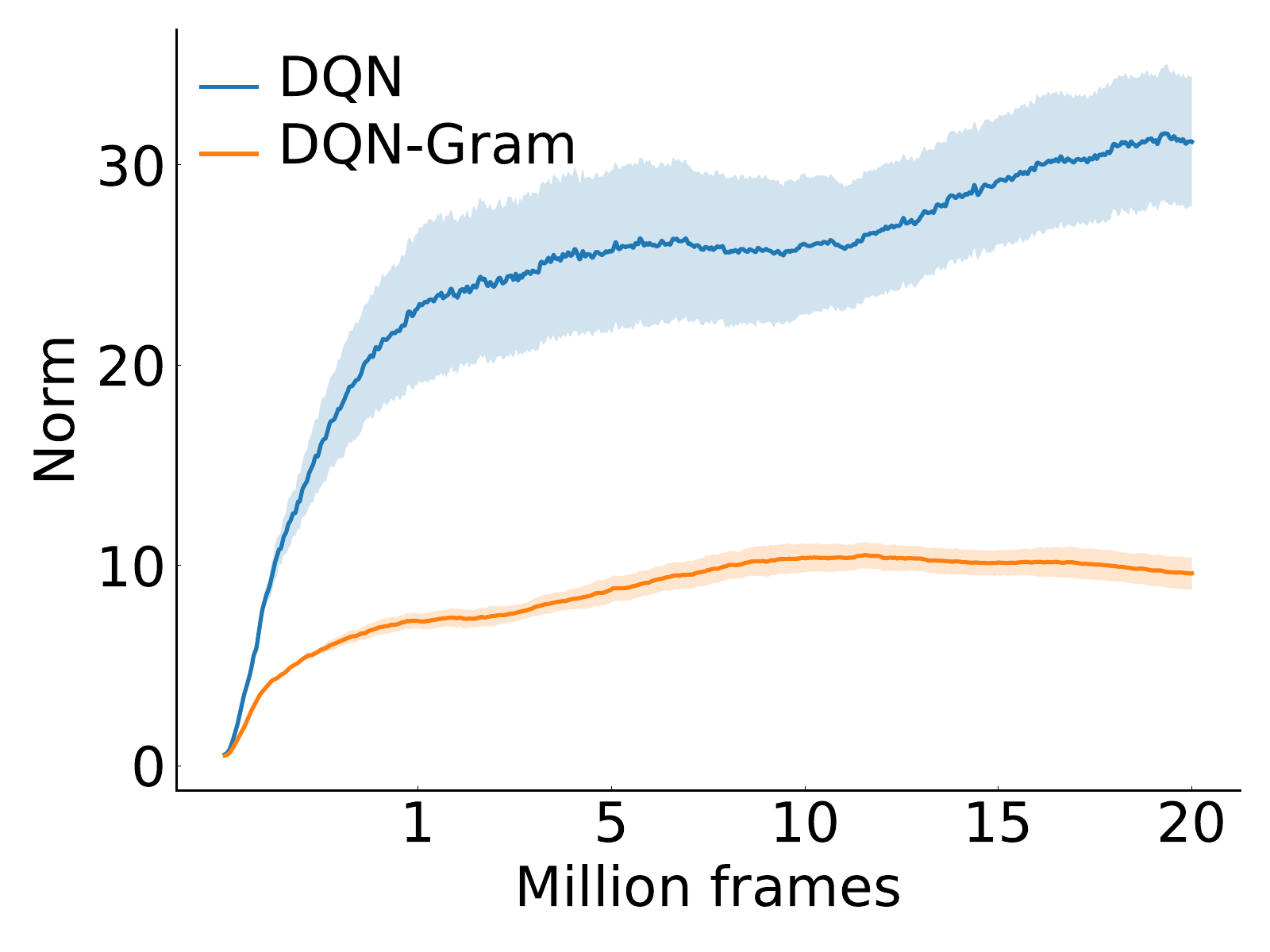}
    \caption{Norm of $\phi$, i.e. last hidden layer activations.}
    \end{subfigure}\qquad
	\begin{subfigure}{.45\linewidth}
        \includegraphics[width=0.9\textwidth]{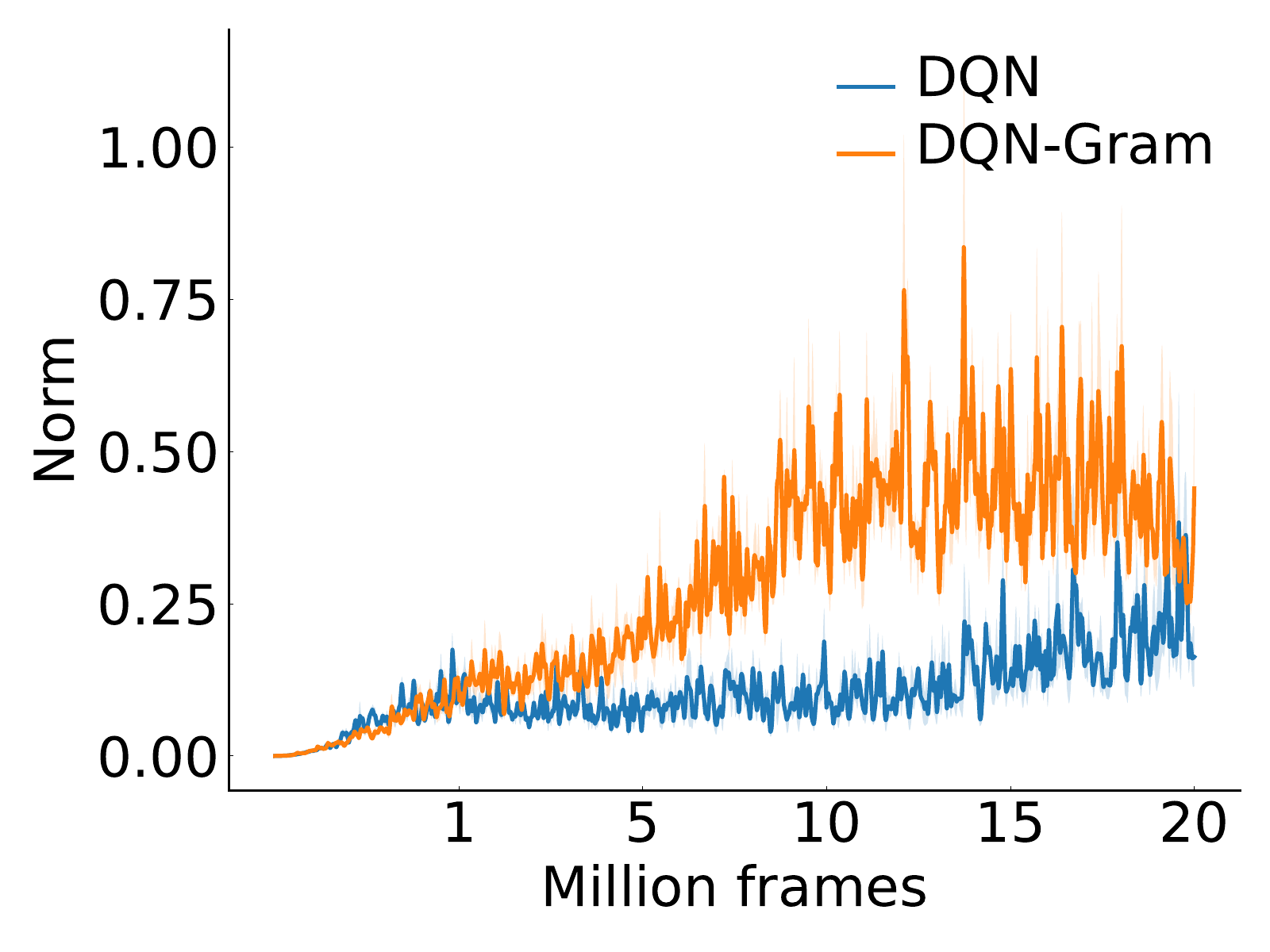}
	\caption{Magnitude of gradients of all layers excluding last linear layer.}
	\end{subfigure}
    \begin{subfigure}{0.45\linewidth}
        \includegraphics[width=0.9\textwidth]{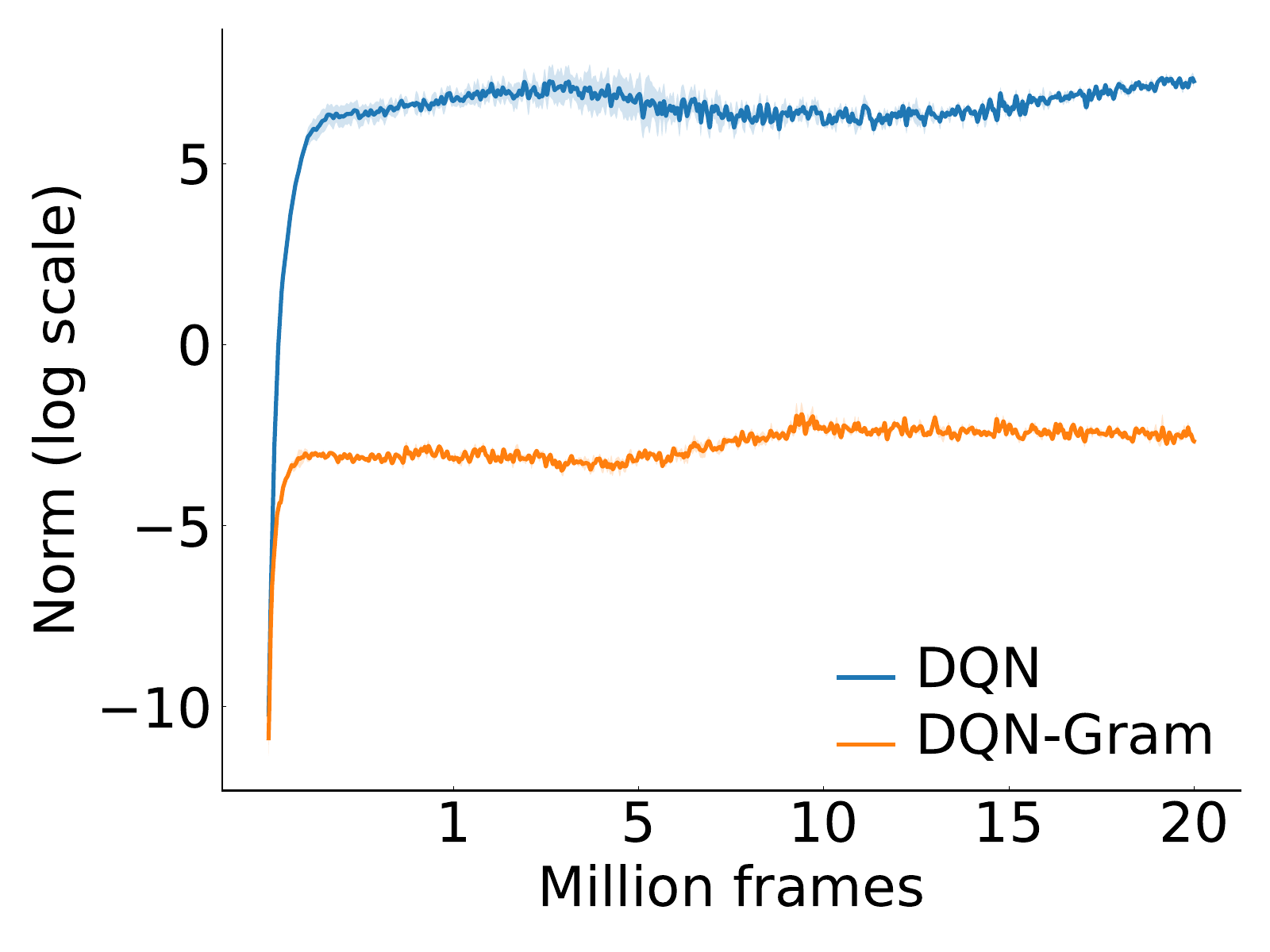}
    \caption{Magnitude of $\phi_{n*}^T\phi_{n*}$}
    \end{subfigure}
	\begin{subfigure}{.45\linewidth}
        \includegraphics[width=0.9\textwidth]{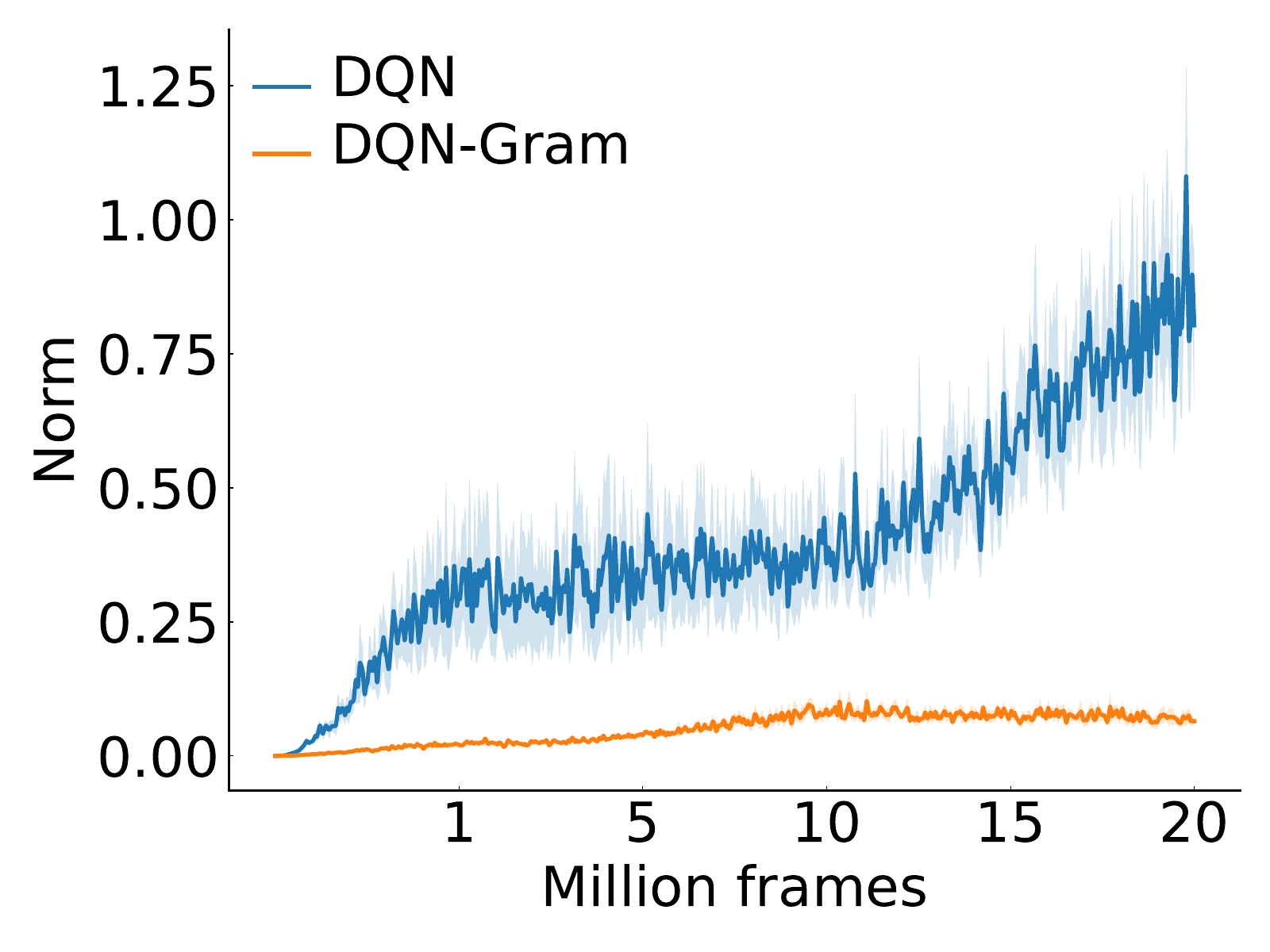}
	\caption{Magnitude of gradients of the last linear layer.}
	\end{subfigure}
    \begin{subfigure}{.45\linewidth}
        \includegraphics[width=0.9\textwidth]{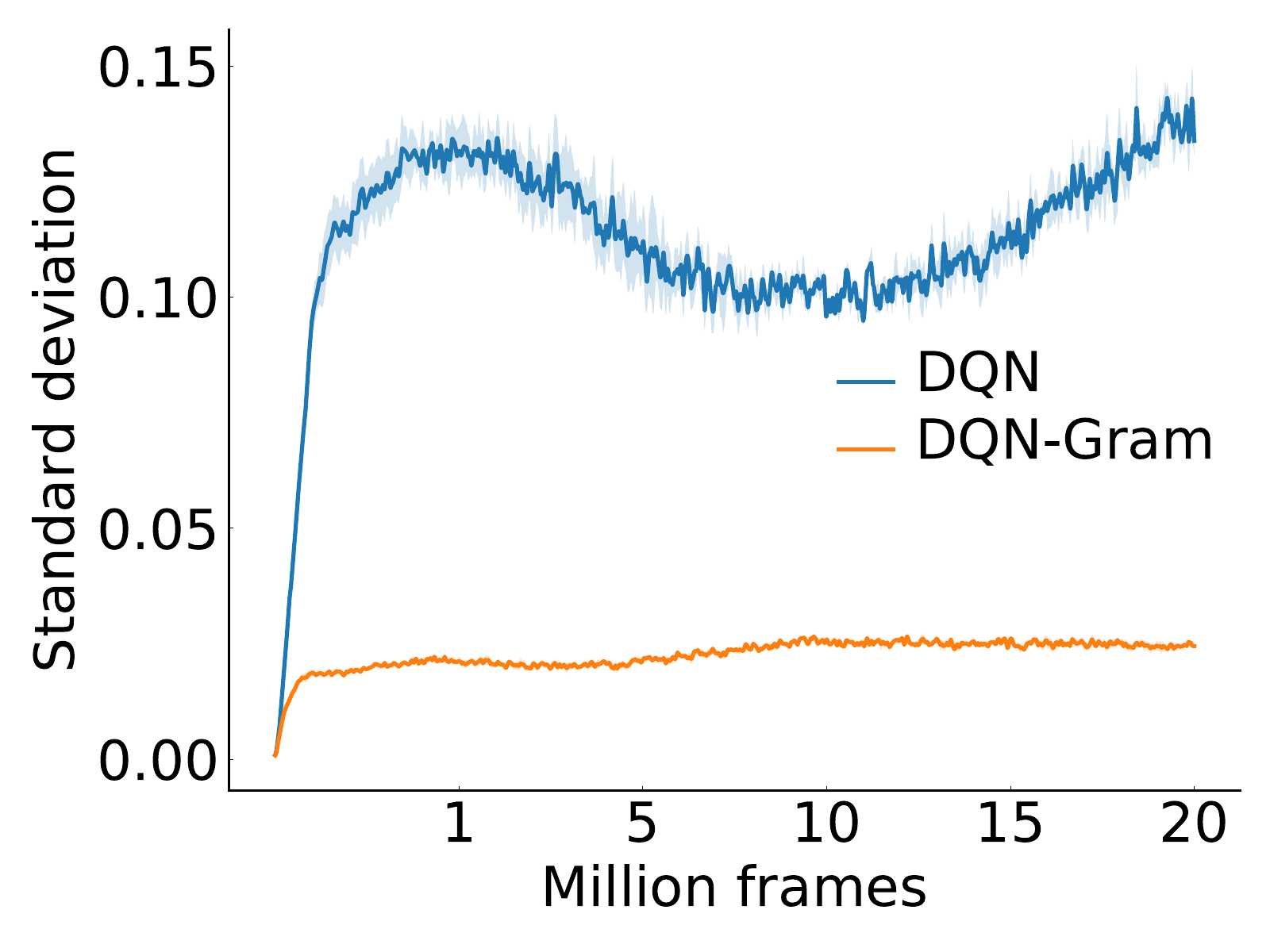}
	\caption{Sum of standard deviations of $\phi_{*d}$}
	\end{subfigure}
	\begin{subfigure}{.45\linewidth}
        \includegraphics[width=0.9\textwidth]{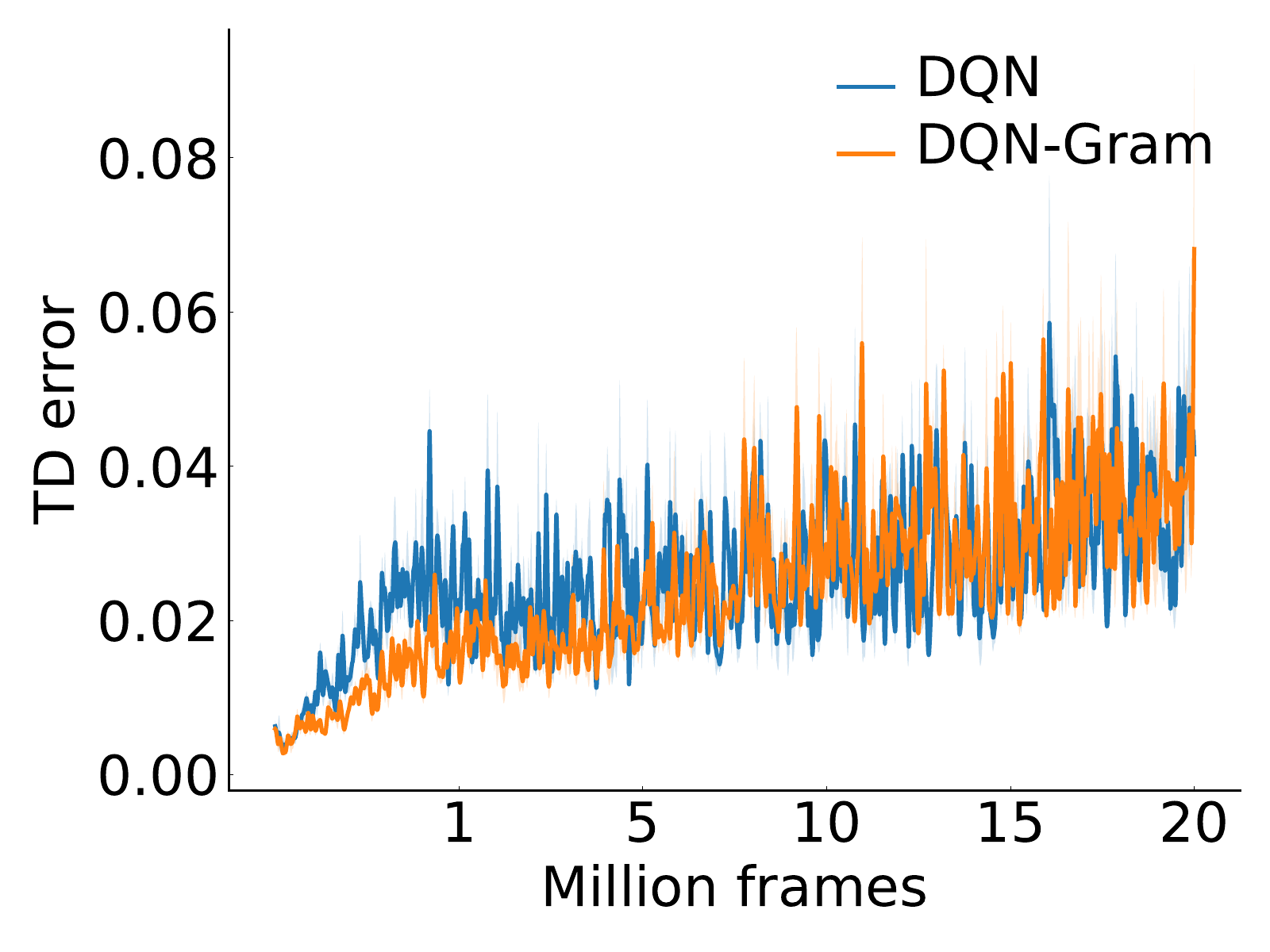}
	\caption{TD error.}
	\end{subfigure}
    \caption{Gradient and feature dynamics during training for Seaquest Atari 2600 averaged over 3 seeds. Bars represent standard errors.}
    \label{fig:grad_norms}
    \end{figure}

Observe that the the Gram regularizer can be decomposed in the following way:
\begin{equation}
	\begin{split}
	& \sum_{n,d}G^2_{n,d} - \sum_d Var(\phi_{*d})^2 = \\
	& \sum_n||\phi_{n*}||^4 + \sum_{n!=m} (\phi_{n*}^T \phi_{m*})^2 - \sum_d Var(\phi_{*d})^2\\
	\end{split}
\end{equation}
Therefore, minimizing the Gram regularizer results in the following:
\begin{itemize}
	\item minimizes the norm of features through $\sum_n||\phi_{n*}||^4$
	\item decorrelates samples through $\sum_{n!=m} (\phi_{n*}^T \phi_{m*})^2$
	\item increases the variance of features via $\sum_d Var(\phi_{*d})^2$
\end{itemize}
where the first dimension enumerated by $n,m$ is the sample number and the second dimension enumerated by $d$ is the feature dimension.
However, these 3 objectives are not independent which in practice introduces trade-offs between them.
For example from Figure \ref{fig:grad_norms} we can see that in the Atari 2600 game of Seaquest the norm of features and sample correlation do drop, but the variance is not changing in the direction of the regularizer: the variance drops despite the pressure being applied by the regularizer to increase it.
In addition, the above decomposition explains the difference in the gradient dynamics introduced by the Gram regularizer. It can be seen from Figure \ref{fig:grad_norms} that the growth of the gradients of the last linear layer in DQN is mainly due to the norm of the features and not the TD error.

\section{Summary}
Representation learning in RL has attracted more attention in recent years due to the advancements in deep learning (DL) and its application in RL.
However, not every approach that improves representation learning in supervised deep learning yields similar gains in RL, e.g. dropout \cite{farebrother2018generalization}.
The main reason of such phenomenon might be that RL differs from supervised learning in its objective. 
Therefore, we introduced a theoretically justifiable regularization approach in RL.
We showed that the feature decorrelating regularizer in RL does not interfere with the main objective and introduced a new algorithm based on it that is proved to converge in policy evaluation.
We showed that our method can be scaled to deep RL in linear computational complexity of the features and quadratic complexity in the mini-batch size.
Finally, we examined the statistical properties of the features in deep RL setting and found that the decorrelating regularizer better exploits the representational capabilities of the neural network by increasing the number of useful features learned.

An area worth investigating in future work is the effect of decorrelating features on generalization to unseen states and similar tasks in RL, which is a significant challenge in RL \cite{farebrother2018generalization}.  Another area of future work is to investigate how decorrelation of features can help improve performance with other enhancements like distributional RL and prioritized sampling in the Rainbow architecture \cite{hessel2018rainbow}. It is likely that the decorrelating regularizer can improve features learned in a similar fashion as distributional RL does if viewed as an auxillary task \cite{bellemare2017distributional}; hence, we think that combining the decorrelating regularizer with Rainbow might be beneficial.

\clearpage
\bibliography{bibliography}
\clearpage
\section{Appendix}

\begin{propm}{1}
	Let $A$ be a global minimum of Equation \eqref{eq:red_aug_loss}. Then $A^T \Phi^T D \Phi A$ is a diagonal matrix. Furthermore, the global minimum values of Equation \eqref{eq:red_aug_loss} and Equation \eqref{eq:full_loss} are equal.
\end{propm}
\begin{proof}
	Let $A$ be a global minimum of Equation \eqref{eq:red_aug_loss}. Assume that $A^T \Phi^T D \Phi A$ is not diagonal. We will derive a contradiction.

	Write $L_{REG}(A) = L_1(A) + L_2(A)$, where $L_2$ is the regularizer term in Equation \eqref{eq:red_aug_loss}, and $L_1$ is the MSVE term. Let $\theta_{TD}^*$ be the global minimum of Equation \eqref{eq:full_loss}. By assumption, $L_2(A) > 0$. Let $V$ by any matrix that diagonalizes $\Phi^T D \Phi$. Such a matrix must exist since $\Phi^T D \Phi$ is real and symmetric. Define $D_\theta := diag(V^T \theta_{TD}^*)$. By definition, we have that $VD_\theta \mathbbm{1} = \theta_{TD}^*$, so that $L_1(VD_\theta) = L_{TD}(\theta_{TD}^*)$. As well, $VD_\theta$ satisfies the following.
	\begin{equation*}
		(VD_\theta)^T \Phi^T D \Phi V D_\theta = D_\theta D D_\theta.
	\end{equation*}

	\noindent Hence, $L_2(VD_\theta) = 0$. We then have
	\begin{equation*}
		\begin{split}
		L_{REG}(VD_\theta) &= L_1(VD_\theta) \\
		 &= L_{TD}(\theta_{TD}^*) \leq L_{TD}(A1)\\
		 &= L_1(A) < L(A).
	\end{split}
	\end{equation*}

	\noindent This is a contradiction by our assumption that $A$ minimizes $L$. Hence, $A^T \Phi^T D \Phi A$ must be diagonal.

	For the second part of the proof, assume that $A$ is a global minimum of Equation \eqref{eq:red_aug_loss} as above. Since from the discussion above we know that $L_2(A) = 0$, we have that $L_{REG}(A) = L_1(A)$. Let $\theta^*_{TD}$ be the global minimum of Equation \eqref{eq:full_loss}. Assume for the sake of contradiction that $L_{TD}(\theta^*_{TD}) < L(A)$. But using the same construction of $VD_\theta$ from above, we would have that $L_{REG}(VD_\theta) < L_{REG}(A)$, which cannot be the case. Hence, $L_{TD}(\theta^*_{TD}) = L_{REG}(A)$.
\end{proof}

\begin{propm}{2}
The projection of a square matrix $A_{n \times n}$ onto the space of orthogonal matrices is given by $UV^T$, where $U$ and $V$ are obtained from the SVD of $A$, $A=U\Sigma V^T$.
\end{propm}
\begin{proof}
The projection of $A$ onto $\mathcal{Q}$ is the solution of $\argmin_{Q^TQ=I} ||A-Q||_F$.
Consider SVD of $A$, $A=U\Sigma V^T$. Then, by the unitary invariance of $||.||_F$
\begin{equation}\label{eq:unitary_invariance}
||A-Q||_F = ||U\Sigma V^T - Q||_F = ||\Sigma - U^TQV||_F.
\end{equation}

Note that since $\mathcal{Q}$ is a group and $U, Q, V \in \mathcal{Q}$, it follows that $U^TQV \in \mathcal{Q}$.
Therefore, $\min_{Q \in \mathcal{Q}}||\Sigma - U^TQV||_F = \min_{Q \in \mathcal{Q}}||\Sigma - Q||_F$.

Taking into account that $\Sigma$ is a diagonal matrix,
	\begin{equation*}
	\begin{split}
	 ||\Sigma - Q||^2_F &= \sum_i(\Sigma_{ii} - Q_{ii})^2 + \sum_{i\ne j}Q^2_{ij} \\
	& = \sum_i (\Sigma^2_{ii} + Q^2_{ii} - 2 \Sigma_{ii} Q_{ii}) + \sum_{i\ne j}Q^2_{ij} \\
	& = \sum_i \Sigma^2_{ii} - 2\sum_i \Sigma_{ii} Q_{ii} + \sum_{i,j}Q^2_{ij} \\
	& = \sum_i \Sigma^2_{ii} - 2\sum_i \Sigma_{ii} Q_{ii} + Tr(Q^TQ) \\
	& = \sum_i \Sigma^2_{ii} - 2\sum_i \Sigma_{ii} Q_{ii} + Tr(I) \\
	& = \sum_i \Sigma^2_{ii} - 2\sum_i \Sigma_{ii} Q_{ii} + n \\
 	\end{split}
	\end{equation*}
Noting that $\Sigma_{ii} \ge 0$ by SVD and $Q\in\mathcal{Q}$,
$$
I = \argmin_{Q^TQ=I} \sum_i \Sigma^2_{ii} - 2\sum_i \Sigma_{ii} Q_{ii} + n
$$
Hence, $\argmin_{Q \in \mathcal{Q}} \|\Sigma - Q\| = U V^T$. The result follows from Equation \eqref{eq:unitary_invariance}.
\end{proof}

\begin{propm}{3}
	$h(A)$ defined by Equation \eqref{eq:red_a_update} over the space of orthogonal matrices satisfies the Lipschitz condition.
\end{propm}
\begin{proof}
	Let $\mathcal{Q} = \{M \in \mathcal{R}^{n \times n} |$ $M$ is orthogonal $\}$. Consider $f: \mathcal{R}^{n \times n} \rightarrow \mathcal{R}^{n \times n}$ defined by $f(M)=M^TM$. Hence, $f(\mathcal{Q})=I$ and $f^{-1}(I)=\mathcal{Q}$. Since $f$ is continuous and $I$ is closed (as a singleton), $f^{-1}(I)=\mathcal{Q}$ is also closed. Note also that $\mathcal{Q}$ is bounded in the operator norm, since for any $Q \in \mathcal{Q}$, $||Q||=sup_x\{||Qx|| : ||x||=1\}=1$. Therefore, $\mathcal{Q}$ is compact by Heine-Borel theorem.

Observe that $h(A)\in \mathcal{C}^\infty$. $h'(A)$ is therefore continuous and reaches its maximum over the compact $\mathcal{Q}$. This means that $h'(A)$ is bounded over $\mathcal{Q}$, so that $h(A)$ satisfies the Lipschitz condition if restricted to $\mathcal{Q}$.
\end{proof}

\begin{propm}{4}
	Assuming that $\sup_n \|A_n \| < \infty$ almost surely and the Robbins-Munro step-size conditions, the iterates of the unprojected stochastic update converge to a compact, connected, internally chain-transitive set of $\dot{A}(t) = h(A(t))$. Additionally, if $\Phi$ is of full rank, there exists such a set that contains at least one equilibrium of $\dot{A}(t) = h(A(t))$.

\end{propm}
\begin{proof}
	The claim follows from satisfying the assumptions for Theorem 2 in \cite[Ch.~2]{borkar2009stochastic}. First, the martingale differences satisfy the required bound by Proposition \ref{prop:martingale_diff}, we assume the Robbins-Munro step-size conditions, and we assume $\sup_n \|A_n\| < \infty$ almost surely. Second, note that the proof that $h$ is Lipschitz in Proposition \ref{prop:lipschitz} still follows through if $\sup_n \|A_n\| < \infty$ and we take $h(A)$ to be defined on a compact ball centered at 0 with radius more than $\sup_n \|A_n\|$.

	The existence of a compact, connected, internally chain-transitive set of $\dot{A}(t) = h(A(t))$ that contains an equilibrium of this ODE follows from Proposition \ref{prop:equilibrium_existence} (namely, the set is the equilibrium point itself).

\end{proof}

\begin{propm}{5}\label{prop:equilibrium_existence}
	Assuming that $\Phi$ is full rank, the following ODE has at least one equilibrium point.
	\begin{equation*}
		\dot{A}(t) = h(A(t)).
	\end{equation*}
\end{propm}
\begin{proof}
	Let $V$ be an orthogonal matrix that diagonalizes $\Phi^T D \Phi$. Let us constrain $A$ to be a diagonal matrix. We aim to solve the following system of equations for $A$.
	\begin{align*}
		0 &= h(VA) = \\
		 &-\sum_s \mu(s) \phi(s) 1^T (R(s) + \gamma \phi(s')^TV A \mathbbm{1} - \phi(s)^T VA \mathbbm{1})\\
		&+ \lambda \sum_{i < j} [e_i  A^T V^T \Phi^T D \phi V A e_j] \Phi^T D \Phi VA[e_j e_i^T e_i e_j^T].
	\end{align*}

	Since $V$ diagonalizes $\Phi^T D \Phi$ and since $A$ is diagonal, we have that $e_i  A^T V^T \Phi^T D \phi V A e_j = 0$ for $i < j$. Hence, the second term in $h(VA)$ vanishes. We are left with
\begin{align*}
		& \sum_{s}\mu(s) \phi(s) \mathbbm{1}^T R(s) \\
		& + \sum_s \mu(s)\phi(s)(\gamma \phi(s')^T  - \phi(s)^T )V A\mathbbm{1}. \\
\end{align*}

	Since we assume that $\Phi$ is full rank and $\gamma < 1$, $\sum_s \mu(s)\phi(s)(\gamma \phi(s')^T  - \phi(s)^T )$ is invertible, as proven for instance in \cite{tsitsiklis1997analysis}. Because $V$ is assumed to be orthogonal and therefore invertible, we can explicitly solve for $A$.
\end{proof}

\subsection{Semi-Gradient of $L_{REG}(A)$}
For generality, let us first derive the semi-gradient of $L_{REG}(\theta, A)$.
\begin{equation} \label{eq:full_theta_update}
\begin{split}
	&\frac{ \partial L(\theta, A)}{ \partial \theta} = \\
	&-\sum_s \mu(s) A^T \phi(s)[R(s) + \gamma \phi(s')^T A \theta - \phi(s)^T A \theta]
\end{split}
\end{equation}
\begin{equation} \label{eq:full_a_update}
\begin{split}
	&\frac{ \partial L(\theta, A)}{ \partial A} = \\
	&-\sum_s \mu(s) \phi(s)\theta^T [R(s) + \gamma \phi(s')^T A \theta - \phi(s)^T A \theta] \\
	&+ \frac{\partial}{\partial A} 0.5 \lambda \sum_{i<j}[e_i\sum_s (\mu(s) A^T \phi(s) \phi(s)^T A) e_j]^2 \\
\end{split}
\end{equation}
\noindent We now expand the second term above.
\begin{equation} \label{eq:full_a_details}
\begin{split}
	&\frac{\partial}{\partial A} 0.5 \lambda \sum_{i<j}[e_i\sum_s (\mu(s) A^T \phi(s) \phi(s)^T A) e_j]^2 \\
	&= \lambda \sum_{i<j} e_i [\sum_s \mu(s) A^T \phi(s) \phi(s)^T A] e_j \\
	& \frac{\partial}{\partial A} [e_i\sum_s \mu(s) A^T \phi(s) \phi(s)^T A e_j] \\
	&= \lambda \sum_{i<j} e_i [\sum_s \mu(s) A^T \phi(s) \phi(s)^T A] e_j \\
	& \sum_s \mu(s) \phi(s) \phi(s)^T A [e_j e_i^T + e_i e_j^T] \\
	&= \lambda \sum_{i<j} [e_i A^T \Phi^T D \Phi A e_j] \Phi^T D \Phi A [e_j e_i^T + e_i e_j^T] \\
	&= \lambda \Phi^T D \Phi A \sum_{i<j} \tilde{\sigma}_{ij}^2[e_j e_i^T + e_i e_j^T] \\
	&= \lambda \Phi^T D \Phi A [\sum_{i<j} \tilde{\sigma}_{ij}^2 e_j e_i^T + \sum_{i<j} \tilde{\sigma}_{ij}^2 e_i e_j^T] \\
	&= \lambda \Phi^T D \Phi A [E_1^T \tilde{D} E_2 + E_2^T \tilde{D} E_1]
\end{split}
\end{equation}
Setting $\theta=\mathbbm{1}$ yields the semi-gradient of $L_{REG}(A)$. %

Define
\begin{equation} \label{eq:em_gradient}
\begin{split}
	&g(A_n) = \\
	&\phi(s) \mathbbm{1}^T [R(s) + \gamma \phi(s')^T A_n \mathbbm{1} - \phi(s)^T A_n \mathbbm{1}]\\
	&+ \lambda \sum_{i<j} [e_i A_n^T \phi \phi^T A_n e_j] \phi \phi^T A_n [e_j e_i^T + e_i e_j^T]\\
\end{split}
\end{equation}
which is the argument of Equation \eqref{eq:red_a_update} at $A_n$. Define also
\begin{equation} \label{eq:martingale_diff}
	M_{n+1} = g(A_n) - \frac{ \partial L(A)}{ \partial A} \Big |_{A_n}
\end{equation}

\begin{propm}{6}\label{prop:martingale_diff}
	Assume that $\sup_n \|A_n\| < \infty$ almost surely, or that we project the iterates $A_n$ to $\mathcal{Q}$ at every step. Then $\{M_n\}$ defined by Equation \eqref{eq:martingale_diff} is
	\begin{itemize}
		\item a martingale difference sequence with respect to
		\begin{equation*}
			\mathcal{F}_n=\sigma(A_m, M_m, m \le n) = \sigma(A_0, M_1, \dots M_n)
		\end{equation*}
		i.e. $\mathbb{E}[M_{n+1}|\mathcal{F}_n]=0$ a.s. $\forall n \in \mathbb{N}$
		\item and the $\{M_n\}$ are square-integrable with
		\begin{equation*}
			\mathbb{E} \big [ ||M_{n+1}||^2 | \mathcal{F}_n \big ] \le K(1 + ||A_n||^2) \hskip 0.1cm \text{a.s.}  \hskip 0.1cm \forall n \in \mathbb{N}
		\end{equation*}
	\end{itemize}
\end{propm}
\begin{proof}
The application of iterated expectations immediately yields $\mathbb{E}[M_{n+1}|\mathcal{F}_n]=0$ a.s. $\forall n \in \mathbb{N}$.

Recall that $\frac{ \partial L(A)}{ \partial A}$ by Proposition \ref{prop:lipschitz}. Also note that $g(A)$ is Lipschitz by the same argument as in Proposition \ref{prop:lipschitz}. Furthermore, if $f$ is Lipschitz, then for any fixed $x_0$,
\begin{equation*}
	\begin{split}
	& ||f(x)|| - ||f(x_0)|| \le ||f(x) - f(x_0)|| \le L ||x - x_0|| \\
	\end{split}
\end{equation*}
Rearranging the terms and applying the triangle inequality one obtains:
\begin{equation} \label{eq:lipschitz2}
	\begin{split}
	& ||f(x)|| \le L ||x - x_0|| + ||f(x_0)|| \\
	& \le L ||x|| - L ||x_0|| + ||f(x_0)|| < K(1 + ||x||)\\
\end{split}
\end{equation}
Hence, by the triangle inequality and Equation \eqref{eq:lipschitz2}:
\begin{equation*}
	\begin{split}
	& ||M_{n+1}|| = \Big| \Big| g(A_n) - \frac{ \partial L(A)}{ \partial A} \Big |_{A_n} \Big| \Big| \\
	& \le ||g(A_n)|| + \Big| \Big|\frac{ \partial L(A)}{ \partial A} \Big |_{A_n} \Big| \Big| \\
	& \le K_1 (1 + ||A_n||) K_2 (1 + ||A_n||) = K (1 + ||A_n||)\\
	\end{split}
\end{equation*}
with $K = K_1 + K_2$.
Therefore,
\begin{equation*}
	\begin{split}
	& ||M_{n+1}||^2 = K^2 (1 + ||A_n||)^2 \\
	& = K^2 (1 + 2 ||A_n||) + ||A_n||^2 \\
	& < K^2 (1 + 2 C + ||A_n||^2) \\
	\end{split}
\end{equation*}
where $C = \sup_{A \in \mathcal{Q}} A \in \mathbb{R}$ by compactness of $\mathcal{Q}$ if we are projecting iterates, or $C = \sup_n \|A_n\| < \infty$ by the other boundedness assumption.
Finally,
\begin{equation*}
	\begin{split}
	& K^2 (1 + 2 C + ||A_n||^2) \\
	& \le K^2 (1 + 2 C) + K^2 ||A_n||^2) \\
	& \le K^2 (1 + 2 C) + K^2 (1 + 2 C) ||A_n||^2) \\
	& K^2 (1 + 2 C) (1 + ||A_n||) = \tilde{K} (1 + ||A_n||^2) \\
	\end{split}
\end{equation*}
Combining last two inequalities, one obtains:
\begin{equation*}
	\begin{split}
	& ||M_{n+1}||^2 \le \tilde{K} (1 + ||A_n||^2) \\
	\end{split}
\end{equation*}
Applying conditional expectations completes the proof.
\end{proof}
\clearpage
\begin{figure*}[ht!]
	\centering
	\includegraphics[width=0.95\textwidth]{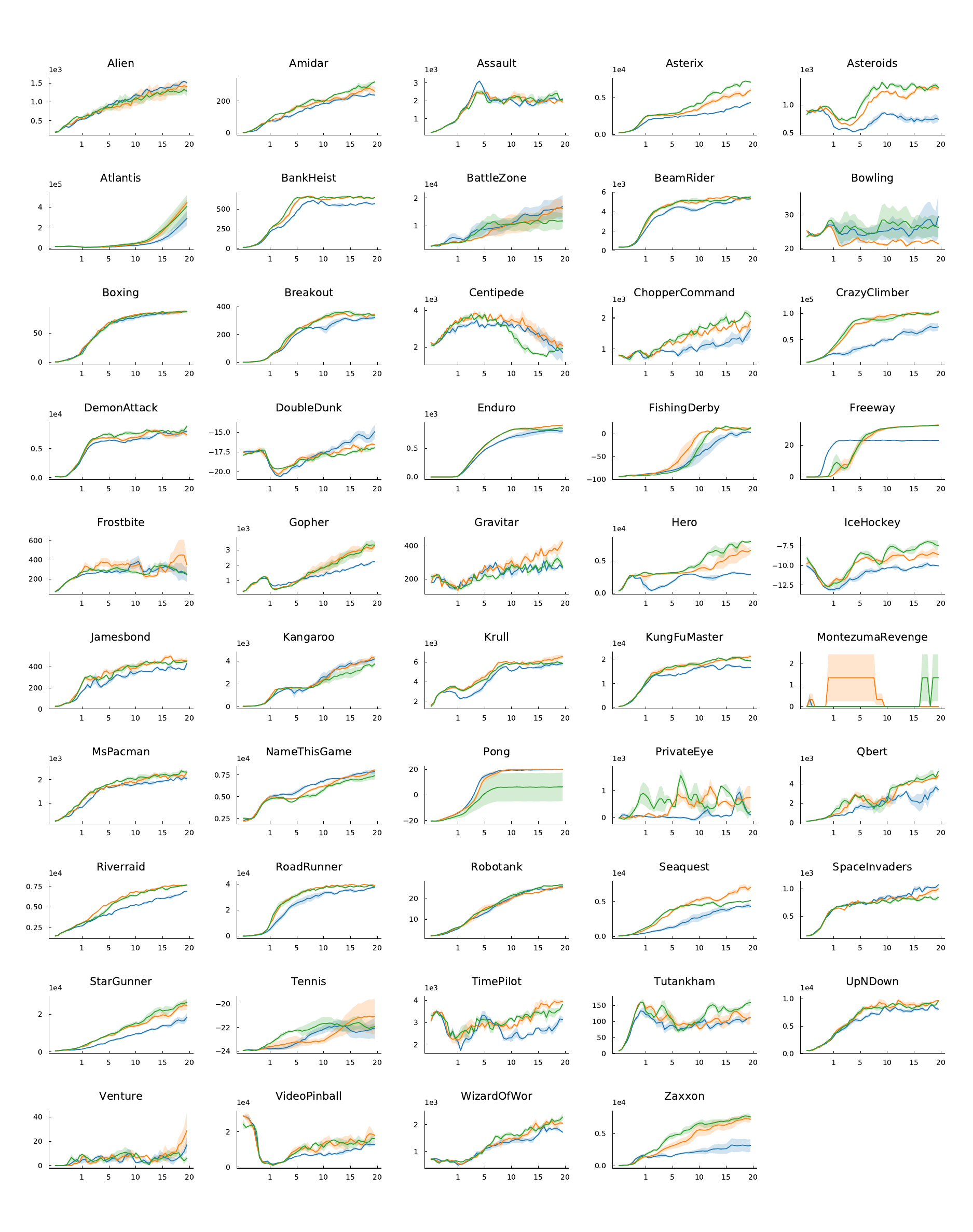}
    \caption{Training curves of \textcolor{green}{DQN-Gram}, \textcolor{orange}{DQN-decor}, \textcolor{blue}{DQN} for 49 Atari 2600 games. Error bars represent standard errors over 3 seeds.}
    \label{fig:all_curves}
\end{figure*}

\clearpage
\begin{table*}[htpb]
\centering  
\begin{tabular}{l|r|r|r|r|r}
\textbf{Game} & \textbf{Random} & \textbf{DQN} & \textbf{DQN-decor} & \textbf{DQN-Gram} \\
\hline
Alien & 227.8 & \textbf{1445.8} & 1348.5 & 1282.2 \\
Amidar & 5.8 & 234.9 & 263.8 & \textbf{292.0} \\
Assault & 222.4 & 2052.6 & 1950.8 & \textbf{2178.8} \\
Asterix & 210.0 & 3880.2 & 5541.7 & \textbf{6851.6} \\
Asteroids & 719.1 & 733.2 & 1292.0 & \textbf{1313.0} \\
Atlantis & 12850.0 & 189008.6 & \textbf{307251.6} & 290744.2 \\
BankHeist & 14.2 & 568.4 & 648.1 & \textbf{650.1} \\
BattleZone & 2360.0 & \textbf{15732.2} & 14945.3 & 11543.1 \\
BeamRider & 363.9 & 5193.1 & 5394.4 & \textbf{5429.7} \\
Bowling & 23.1 & \textbf{27.3} & 21.9 & 26.4 \\
Boxing & 0.1 & 85.6 & 86.0 & \textbf{87.2} \\
Breakout & 1.7 & 311.3 & 337.7 & \textbf{344.7} \\
Centipede & 2091.0 & 2161.2 & \textbf{2360.5} & 1721.2 \\
ChopperCommand & 811.0 & 1362.4 & 1735.2 & \textbf{1976.7} \\
CrazyClimber & 10781.0 & 69023.8 & \textbf{100318.4} & 99903.6 \\
DemonAttack & 152.1 & 7679.6 & 7471.8 & \textbf{8081.3} \\
DoubleDunk & -18.6 & \textbf{-15.5} & -16.8 & -17.2 \\
Enduro & 0.0 & 808.3 & \textbf{891.7} & 839.0 \\
FishingDerby & -91.7 & 0.7 & 11.7 & \textbf{12.3} \\
Freeway & 0.0 & 23.0 & \textbf{32.4} & 32.3 \\
Frostbite & 65.2 & 293.8 & \textbf{376.6} & 295.4 \\
Gopher & 257.6 & 2064.5 & 3067.6 & \textbf{3073.2} \\
Gravitar & 173.0 & 271.2 & \textbf{382.3} & 295.4 \\
Hero & 1027.0 & 3025.4 & 6197.1 & \textbf{7687.6} \\
IceHockey & -11.2 & -10.0 & -8.6 & \textbf{-7.3} \\
Jamesbond & 29.0 & 387.5 & \textbf{471.0} & 443.6 \\
Kangaroo & 52.0 & 3933.3 & \textbf{3955.5} & 3363.3 \\
Krull & 1598.0 & 5709.9 & \textbf{6286.4} & 5912.5 \\
KungFuMaster & 258.5 & 16999.0 & \textbf{20482.9} & 20121.1 \\
MontezumaRevenge & 0.0 & 0.0 & 0.0 & \textbf{0.8} \\
MsPacman & 307.3 & 2019.0 & 2166.0 & \textbf{2250.3} \\
NameThisGame & 2292.0 & \textbf{7699.0} & 7578.2 & 7112.9 \\
Pong & -20.7 & 19.9 & \textbf{20.0} & 6.2 \\
PrivateEye & 24.9 & 345.6 & \textbf{610.8} & 551.8 \\
Qbert & 163.9 & 2823.5 & 4432.4 & \textbf{4750.7} \\
Riverraid & 1339.0 & 6431.3 & \textbf{7613.8} & 7446.1 \\
RoadRunner & 11.5 & 35898.6 & \textbf{39327.0} & 38306.9 \\
Robotank & 2.2 & 24.8 & 24.5 & \textbf{26.0} \\
Seaquest & 68.4 & 4216.6 & \textbf{6635.7} & 4965.3 \\
SpaceInvaders & 148.0 & \textbf{1015.8} & 913.0 & 820.1 \\
StarGunner & 664.0 & 15586.6 & 21825.0 & \textbf{23870.5} \\
Tennis & -23.8 & -22.3 & \textbf{-21.2} & -21.8 \\
TimePilot & 3568.0 & 2802.8 & \textbf{3852.1} & 3494.0 \\
Tutankham & 11.4 & 103.4 & 116.2 & \textbf{142.7} \\
UpNDown & 533.4 & 8234.5 & \textbf{9105.8} & 9050.8 \\
Venture & 0.0 & 8.4 & \textbf{15.3} & 7.5 \\
VideoPinball & \textbf{16257.0} & 11564.1 & 15759.3 & 14627.2 \\
WizardOfWor & 563.5 & 1804.3 & 2030.3 & \textbf{2081.5} \\
Zaxxon & 32.5 & 3105.2 & 7049.4 & \textbf{7459.4} \\
\end{tabular}
\caption{Final training performance for 49 Atari 2600 games averaged over 3 seeds.}
\end{table*}

\bibliographystyle{aaai}
\end{document}